\documentclass[11pt]{article}

\usepackage[preprint]{acl}

\usepackage{times}
\usepackage{latexsym}

\usepackage[T1]{fontenc}

\usepackage[utf8]{inputenc}

\usepackage{microtype}

\usepackage{inconsolata}

\usepackage{graphicx}

\usepackage{enumitem}
\setlist{leftmargin=*}
\usepackage{amsmath} 
\usepackage{booktabs}
\usepackage{graphicx}
\usepackage{enumitem}
\usepackage{tabularx}
\usepackage{makecell}
\usepackage{cleveref}
\usepackage[table]{xcolor}
\usepackage{multicol}
\usepackage{tcolorbox}
\tcbuselibrary{breakable}
\tcbuselibrary{skins}
\usepackage{xcolor}
\usepackage{longtable}
\usepackage{array}
\usepackage{ragged2e}
\usepackage{amssymb}
\usepackage{float}
\usepackage{minted}
\usepackage{url}
\usepackage{hyperref}
\usepackage{cuted}

\definecolor{mintgreen}{RGB}{220, 245, 230}

\title{\textsc{UltraLogic}: Enhancing LLM Reasoning through Large-Scale Data Synthesis and Bipolar Float Reward}

\author{First Author \\
  Affiliation / Address line 1 \\
  Affiliation / Address line 2 \\
  Affiliation / Address line 3 \\
  \texttt{email@domain} \\\And
  Second Author \\
  Affiliation / Address line 1 \\
  Affiliation / Address line 2 \\
  Affiliation / Address line 3 \\
  \texttt{email@domain} \\}

\author{
    \textbf{Yile Liu}$^{1,2}$\thanks{Equal contribution.}, 
    \textbf{Yixian Liu}$^{1}$\footnotemark[1]\thanks{Corresponding author.}, 
    \textbf{Zongwei Li}$^{1}$\footnotemark[1], 
    \textbf{Yufei Huang}$^{1}$, 
    \textbf{Xinhua Feng}$^{1}$, 
    \textbf{Zhichao Hu}$^{1}$ \\
    \textbf{Jinglu Hu}$^{2}$, 
    \textbf{Jianfeng Yan}$^{1}$, 
    \textbf{Fengzong Lian}$^{1}$, 
    \textbf{Yuhong Liu}$^{1}$ \\
    \textsuperscript{1}Hunyuan, Tencent \\
    \textsuperscript{2}Waseda University \\
    \texttt{irei.liu@asagi.waseda.jp, ly\_xian@whu.edu.cn, zongweili@tencent.com} \\
}

\begin{document}
\maketitle
\begin{abstract}

While Large Language Models (LLMs) have demonstrated significant potential in natural language processing , complex general-purpose reasoning—requiring multi-step logic, planning, and verification—remains a critical bottleneck. Although Reinforcement Learning with Verifiable Rewards (RLVR) has succeeded in specific domains , the field lacks large-scale, high-quality, and difficulty-calibrated data for general reasoning. To address this, we propose \textsc{UltraLogic}, a framework that decouples the logical core of a problem from its natural language expression through a Code-based Solving methodology to automate high-quality data production. The framework comprises hundreds of unique task types and an automated calibration pipeline across ten difficulty levels. Furthermore, to mitigate binary reward sparsity and the Non-negative Reward Trap, we introduce the Bipolar Float Reward (BFR) mechanism, utilizing graded penalties to effectively distinguish perfect responses from those with logical flaws. Our experiments demonstrate that task diversity is the primary driver for reasoning enhancement , and that BFR, combined with a difficulty matching strategy, significantly improves training efficiency, guiding models toward global logical optima.
\end{abstract}

\section{Introduction}

In recent years, Large Language Models (LLMs) have achieved revolutionary breakthroughs in numerous areas of Natural Language Processing (NLP)~\cite{achiam2023gpt, deepseekai2025deepseekv3technicalreport, yang2025qwen3technicalreport}. However, complex reasoning—particularly general-purpose reasoning that requires multi-step logic, planning, and verification—remains a critical bottleneck in advancing model intelligence to higher levels~\cite{liu2025agenticmathenhancingllmreasoning, wu-etal-2025-c2rbench}. To overcome this limitation, both academia and industry have turned their attention to the post-training stage, especially methods based on Reinforcement Learning (RL)~\cite{ouyang2022traininglanguagemodelsfollow, rafailov2024directpreferenceoptimizationlanguage}. These models have demonstrated astonishing improvements in reasoning abilities in domains like mathematics and code by leveraging Reinforcement Learning with Verifiable Rewards (RLVR)~\cite{shao2024deepseekmathpushinglimitsmathematical}. The core of this paradigm lies in the fact that task answers in these domains (e.g., whether code passes unit tests or a mathematical answer is correct) have explicit, automatically verifiable feedback, providing a clear reward signal for reinforcement learning.

However, when extending this paradigm from specific domains to broader, more general-purpose reasoning tasks, a fundamental problem becomes particularly prominent: the insufficient supply of high-quality training data~\cite{liu2025synlogicsynthesizingverifiablereasoning, lu2024mathgeniegeneratingsyntheticdata, liu2025agenticmathenhancingllmreasoning}. The success of RLVR currently relies heavily on existing high-quality competition datasets~\cite{hendrycks2021measuringmathematicalproblemsolving, rein2023gpqagraduatelevelgoogleproofqa}, but the field of general reasoning lacks similar large-scale data resources. Existing datasets are not only limited in task diversity~\cite{liu2025synlogicsynthesizingverifiablereasoning, li202512surveyreasoning}, failing to cover a wide range of reasoning scenarios, but they also generally lack a clear and controllable difficulty calibration system~\cite{kwan2025opensir, wang2025morphobench}. This makes the difficulty distribution of the training data hard to manage, which in turn affects the model's learning efficiency and stability.

To systematically address this data supply problem, we propose and implement \textbf{\textsc{UltraLogic}}: an innovative framework for the large-scale, automated production of high-quality reasoning data, built upon a core Code-based Solving Framework methodology. This framework operates through a process that combines human definition with automated production. First, in a collaborative process between domain experts and prompt-driven LLMs, two critical Python functions are written for each novel reasoning task type: an input function to generate the slot-filling data for predefined problem templates, and a solution function to provide a deterministic ground-truth answer based on that data. This step ensures the logical correctness, verifiability, and controllability of all our data. Subsequently, an automated pipeline takes over, performing large-scale generalization on these "seed tasks" using Programmatic Expansion (PE) techniques~\cite{mishra2023lilaunifiedbenchmarkmathematical}. This pipeline not only generates a vast number of problem variants but also systematically creates task instances spanning 10 difficulty levels by precisely controlling the parameters of the input function. These difficulty levels are then objectively calibrated against the measured success rates of flagship model. Experiments show that by training on our synthesized data alone, using a standard binary reward mechanism~\cite{shao2024deepseekmathpushinglimitsmathematical}, the model exhibits a significant improvement in general reasoning capabilities compared to baselines.

Building on this foundation, we further explore potential methods for improving training efficiency, which constitutes our second exploratory contribution. We hypothesize that although the binary reward mechanism is clear and effective, providing the learning process with a more information-dense and graded reward signal could potentially enhance both training efficiency and final performance. To this end, we design and implement a novel \textbf{Bipolar Float Reward (BFR)} mechanism, which introduces a penalty-driven optimization signal to the reinforcement learning process. This mechanism is capable of quantifying "partially correct" outputs based on task-specific characteristics, using metrics such as accuracy or F1-Score~\cite{paulus2017deepreinforcedmodelabstractive}. Our experiments demonstrate that this bipolar approach significantly outperforms both binary and standard non-negative float rewards, facilitating faster convergence and superior final performance by effectively penalizing imperfect reasoning paths.

In summary, the contributions of this paper are twofold: first, we provide an automated framework for the large-scale production of diverse, difficulty-calibrated reasoning data, demonstrating that task diversity is a more critical driver for general reasoning enhancement than mere data scaling; second, we propose the Bipolar Float Reward (BFR) mechanism and provide empirical evidence that integrating a graded penalty into the reward signal is crucial for breaking performance bottlenecks in complex reasoning tasks.

\section{Related Work}

\paragraph{Data Synthesis for Logic and Reasoning.}
Data synthesis for logical reasoning tasks currently follows two primary paradigms: 
programmatic synthesis, which uses deterministic generators to produce data and answers~\cite{liu2025synlogicsynthesizingverifiablereasoning}; and generative synthesis, which leverages powerful LLMs to generate question-answer pairs~\cite{liu2025agenticmathenhancingllmreasoning, yu2024metamathbootstrapmathematicalquestions, lu2024mathgeniegeneratingsyntheticdata}. Our \textsc{UltraLogic} framework combines these two approaches, ensuring the quantity, quality, and diversity of the generated problems while also guaranteeing the reproducibility and verifiability of the generation pipeline. 

\paragraph{Fine-Grained Reward Mechanisms.}
In reinforcement learning, the standard binary (0/1) reward signals are often "overly sparse and lacking in discrimination", hindering learning efficiency. Consequently, the field has shifted toward Process Reward Models (PRMs) \cite{lightman2023letsverifystepstep} which reward each reasoning step rather than just final answers \cite{ma2023letsrewardstepstep}. To mitigate the high human annotation costs of PRMs, "automated PRMs" like OpenPRM \cite{zhang2025openprm} and DG-PRM \cite{yin2025dynamicgeneralizableprocessreward} derive granular signals at the cost of an ORM. Similarly, \textsc{UltraLogic}'s BFR uses pre-defined criteria to reverse-engineer answers for "process-level scoring" without accessing model reasoning traces, offering a low-cost, high-density "middle-ground" solution.

\begin{figure*}[t]
    \centering
    \includegraphics[width=\textwidth]{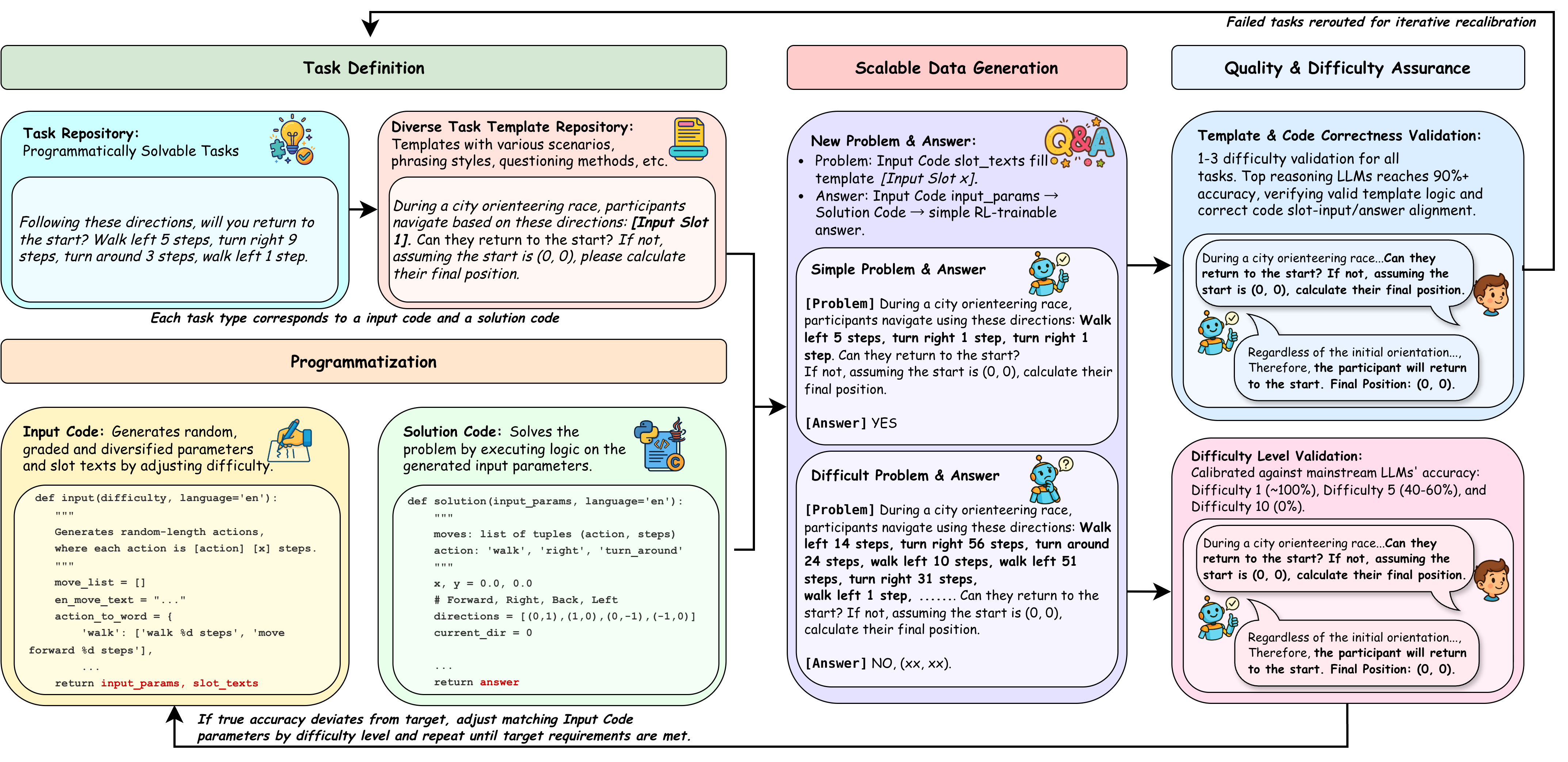} 
    \caption{The overall architecture of the \textsc{UltraLogic} Data framework.}
    \label{fig:architecture}
\end{figure*}

\paragraph{Role of Difficulty in Reinforcement Learning.} 
Problem difficulty is a key variable for LLM training efficiency and performance. Works like OpenSIR \cite{kwan2025opensir} and MorphoBench \cite{wang2025morphobench} utilize difficulty to guide training or adjust evaluation dynamically. Given the sensitivity to difficulty, curriculum learning has become essential, with frameworks such as E3-RL4LLMs \cite{liao2025enhancingefficiencyexplorationreinforcement} and SEELE \cite{li2025stayingsweetspotresponsive} internalizing dynamic adjustments to maintain "high-efficiency regions." Our \textsc{UltraLogic} addresses this via an automated 1–10 calibrated difficulty ladder. We further identify a "Difficulty Matching Phenomenon," proving RL is most effective within the "Zone of Proximal Development" where task difficulty aligns with model capacity.

\section{The \textsc{UltraLogic} Data Framework}

To systematically address the problem of insufficient high-quality training data in the general reasoning domain, we have designed and implemented an innovative framework for the large-scale, automated production of high-quality reasoning data. The core of this framework lies in its design for data diversity, controllable difficulty, and production scalability. This chapter will detail its overall architecture, data synthesis pipeline, and difficulty calibration system.

\subsection{Overall Architecture}
The core methodology of \textsc{UltraLogic} is a system we term the "Code-based Solving Framework." The design philosophy of this framework is to decouple the \textit{logical core} of a reasoning problem from its \textit{natural language expression}. This approach allows us to independently and programmatically control the intrinsic complexity and the external presentation of a problem, thereby enabling the production of diverse data at scale.

As shown in Figure~\ref{fig:architecture}, the \textsc{UltraLogic} framework is primarily composed of three core static components: an Original Task Repository, a Diverse Task Template Repository, and the Data Synthesis Pipeline which acts as the core engine. The Data Synthesis Pipeline itself consists of three key modules: Input Code, Solution Code, and a Difficulty Control Module. The entire workflow begins by selecting a task from the repository and matching it with a template. The pipeline then generates a concrete problem instance complete with a deterministic answer and an objective difficulty rating.


\subsection{Original Task Repository}
\label{Original Task Repository}
The cornerstone of \textsc{UltraLogic}'s diversity is a repository containing hundreds of unique task types. To ensure a systematic and novel coverage of reasoning skills, we developed a three-dimensional orthogonal classification system. Each task is uniquely situated along three dimensions: Task Domain (defining the problem context), Core Reasoning Ability (characterizing the cognitive focal point), and Difficulty Setup (dissecting the source of complexity). We constructed a massive question bank from these types and performed rigorous data cleaning to exclude tasks similar to existing benchmarks. Detailed taxonomies and category definitions are provided in~\Cref{sec:Three-dimensional Orthogonal Categories}. The detailed task category cases are provided in~\Cref{Task Category Cases}.

\subsection{Diverse Task Template Repository}
To transform the abstract logical cores into human-readable natural language problems and to prevent the model from learning fixed textual patterns, we have built a diverse template repository for each task type. The construction of this repository is a semi-automated workflow, accomplished through a collaboration between prompt-driven LLMs and annotators.

This process involves two main stages. The first stage is \textbf{template deconstruction and abstraction}. We begin by providing a concrete, original reasoning problem as input and guide an LLM with a prompt (detailed in~\Cref{The Prompt of Task Mining and Template Creation}) to analyze it. The prompt instructs the model to identify the variable parameters (e.g., specific numbers, names, or operation sequences) and the invariant core logic of the problem. Based on this analysis, the LLM automatically generates a base template with clearly defined "slots", such as \textit{initial state} and \textit{operation sequence}. An annotator then reviews and refines this machine-generated template to ensure its logical rigor and the accuracy of the slot definitions. Furthermore, our task repository is enriched by logic structures inspired by competitive programming platforms such as LeetCode. Unlike natural language puzzles, these tasks possess an intrinsic logical structure that eliminates the need for LLM-driven deconstruction, while offering the unique advantage of verified solution code and deterministic ground-truth answers. The second stage is \textbf{template expansion and enrichment}. After obtaining a human-verified base template, we again leverage an LLM for creative scenario-packing (detailed prompt is provided in~\Cref{The Prompt of Question Template Generalization}). We feed the base template to the LLM with another prompt, instructing it to generate multiple (e.g., 10) template variants with different narrative backgrounds--such as logistics scenarios, spy stories, or sci-fi settings--while keeping the core logic and slots unchanged. Finally, the annotator once again filters and polishes this batch of generated templates, assessing their diversity, linguistic fluency, and logical consistency, to finalize the standardized template library for the task type.

The template repository has native support for both Chinese and English, with templates provided for each language to support bilingual model training.

\subsection{Data Synthesis Pipeline}
\subsubsection{Input Code Generation}

Before implementation, we first employ an LLM-based feasibility assessment to determine whether a reasoning task is suitable for programmatic generation. For tasks that pass this check, we define a core input code, implemented as an \texttt{input} function. The creation of this function is a semi-automated process designed for scalability: the task logic and structural parameters identified during the template deconstruction phase serve as the specification; a prompt-driven LLM then generates the full Python function based on this specification (detailed prompt is provided in~\Cref{The Prompt of Input Code Generation}); and the annotator's final role is to review, debug, and calibrate this machine-generated code to ensure it correctly produces parameters according to the difficulty requirements.

The primary responsibility of the final \texttt{input(difficulty, language)} function is to programmatically generate the core parameters (or ``slot-filling data'') for populating the template slots. It takes a difficulty level \texttt{difficulty} and a target language \texttt{language} as input and contains a set of rules for generating parameters related to the intrinsic logic of the task. Its output is not a complete problem, but rather a structured set of data that constitutes the unique logical core of that problem instance. Crucially, since these parameters are generated stochastically within defined constraints, the framework can theoretically produce an infinite supply of unique reasoning problems for any given task type.
\subsubsection{Solution Code Generation}
To ensure that every piece of generated data has an absolutely correct and verifiable answer, we equip each task type with solution code, implemented as a \texttt{solution} function. Similar to the input code, its creation is also semi-automated. Given the logic of a task, an LLM generates a candidate solution function (The detailed prompt is provided in~\Cref{The Prompt of Solution code Generation}). An annotator then rigorously verifies the correctness and efficiency of this code, ensuring it can reliably solve any problem instance generated by the corresponding \texttt{input} function.

The final \texttt{solution(params, language)} function receives the exact same core parameters \texttt{params} generated by the \texttt{input} function as its input. It implements a deterministic algorithm or set of rules to solve the problem. Because the \texttt{solution} and \texttt{input} functions share the same set of parameters as the basis for problem generation and solving, we guarantee by design a perfect synchronization between "problem" and "answer" to ensure accuracy. 
\subsubsection{Difficulty Control Module}

The Difficulty Control Module is central to \textsc{UltraLogic}'s ability to provide fine-grained, objective grading through a unified 1--10 difficulty ladder. To ensure settings are objective and reproducible, we implement a closed-loop automated calibration process. First, we predefine target success rates $P(d)$ for specific levels (e.g., approximately 100\%, 70\%, 50\%, 30\%, and 0\% for levels 1, 3, 5, 7, and 10, respectively). The system then generates test samples via the \texttt{input} and \texttt{solution} functions and calculates the actual average success rate $P_{\text{actual}}$ using multiple open-source flagship models. If a deviation exists, an automated algorithm dynamically adjusts the internal complexity parameters within the \texttt{input} and \texttt{solution} functions such as increasing reasoning steps or adding constraints until $P_{\text{actual}}$ converges to the target margin. This iterative process essentially follows the ReAct paradigm to achieve precise difficulty alignment~\cite{yao2022react}. Once the parameter configurations are solidified, this programmatic approach allows for unlimited scalability, enabling the framework to adapt to and challenge future models by extending difficulty definitions beyond the current scale. The prompt guiding this automated adjustment is provided in~\Cref{The Prompt of Dynamic Parameter Adjustment}.

\subsubsection{Data Validation and Quality Assurance}
\label{Data Validation and Quality Assurance}
Before full-scale dataset production, each task type and its associated templates undergo a rigorous validation process to ensure the synergy between programmatic logic and linguistic expression. We generate a representative sample set covering every template variant at the lowest difficulty tiers (Levels 1--3) and evaluate them using a flagship model to verify the textual logic and linguistic correctness of each template. A task is deemed valid only if the target rate reaches the predefined target success threshold (above 90\%). This verification step acts as a final quality gate, confirming that the natural language templates correctly convey the constraints generated by the \texttt{input} code and that the \texttt{solution} functions remain robust under various parameterizations. Only tasks and templates that successfully pass this stage proceed to the final training set construction.

A complete metadata example of a task instance is provided in~\Cref{A Complete Task Case}.

\section{Bipolar Float Rewards}

To mitigate the sparsity and poor discrimination of binary rewards, we introduce the Bipolar Float Reward (BFR) mechanism. This chapter details the BFR design, utilizing dense graded feedback to overcome training bottlenecks in multi-step reasoning.

\subsection{Baseline Training Paradigm and the Limitations of Binary Rewards}
We adopt Reinforcement Learning with Verifiable Rewards (RLVR) as our core training paradigm and choose Group Relative Policy Optimization (GRPO)~\cite{shao2024deepseekmathpushinglimitsmathematical} as our primary policy optimization algorithm. In our baseline experiments, we initially used a standard binary reward function: the reward is 1 only if the final answer generated by the model is identical to the ground truth, and 0 otherwise. This clear and explicit reward mechanism serves as a solid foundation for validating the intrinsic value of our \textsc{UltraLogic} dataset.

However, when dealing with the tasks generated by \textsc{UltraLogic}, which include numerous complex, multi-step reasoning problems, we observe that the binary reward has a significant limitation: the reward signal is overly sparse and lacks discrimination. When the model's answer is close to the ground truth, it may indicate that the model has completed most of the reasoning correctly, but due to minor errors in the process, the reward it receives (0) is identical to that of a reasoning path that was completely wrong from the start. This mechanism cannot provide the model with any information about the "degree of correctness," leading to a waste of valuable, near-correct exploratory actions, which can potentially reduce learning efficiency.

\subsection{Design and Iteration of the Graded Float Reward}
To overcome the limitations of binary rewards, we hypothesize that a more information-dense, graded reward signal that reflects the "degree of correctness" could provide the model with richer and more fine-grained learning guidance, thereby accelerating convergence and improving final performance. Based on this motivation, we designed and implemented a novel Graded Float Reward mechanism.

\begin{figure*}[t]  
    \centering
\includegraphics[width=0.75\textwidth]{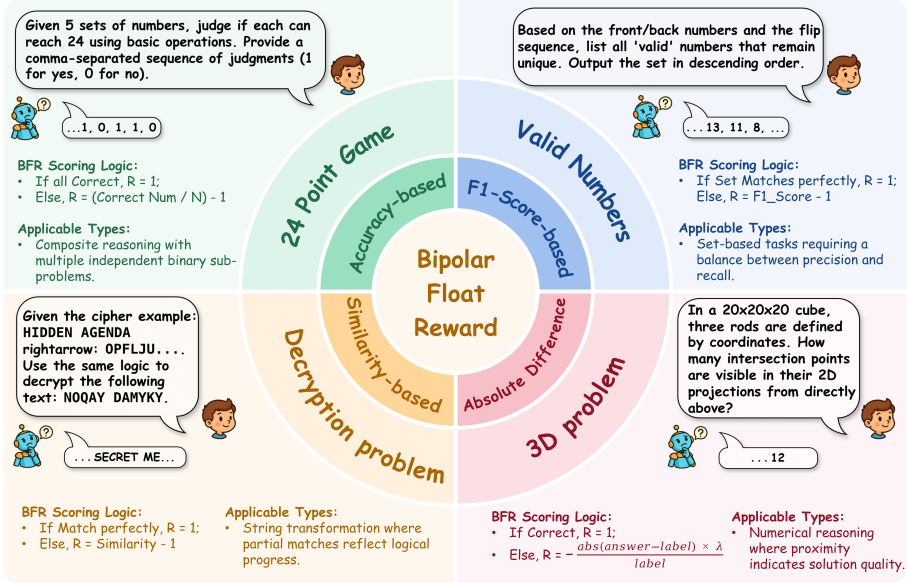}
    \caption{Four scoring methods of the BFR mechanism and illustrative examples. Each task type is matched with the most appropriate scoring method to effectively quantify the correctness of model responses based on its unique characteristics, mapping partially correct answers into graded negative penalties.}
    \label{fig:graded_example}
\end{figure*}

\subsubsection{Initial Exploration: Graded Float Reward in the [0, 1] Range}

Our initial exploration extended rewards from the discrete $\{0, 1\}$ set to the continuous $[0, 1]$ interval to quantify partial correctness. By converting single answers into structured, high-entropy outputs as proxies for reasoning processes, we implemented four task-specific scoring metrics: Accuracy, F1-score, Similarity, and Absolute Difference Rate (see~\Cref{fig:graded_example} for detailed logic and examples). Although this approach provided more granular feedback than binary rewards, preliminary experiments showed that models often plateaued at sub-optimal solutions. We hypothesize that providing non-zero rewards for any flawed response—as long as it is not the worst possible—reduces the model's incentive to explore perfect logic chains, leading it to "settle for" partially correct answers.

\subsubsection{Bipolar Float Reward: Graded Penalties}
\label{sec:bfr_logic}
Based on the observations above, we performed a key iteration. We hypothesize that for imperfect answers, not only should they not receive a positive reward, but they should also be penalized with varying intensity according to their degree of error. To this end, we adjusted the reward range from $[0, 1]$ to $[-1, 0) \cup \{1\}$. The mathematical logic behind this design is rooted in the optimization mechanics of GRPO.

\paragraph{Reshaping Advantage and the Non-negative reward trap.}
In the GRPO framework, the advantage function $\hat{A}_{i,g}$ measures how much better a specific sample is compared to the group average. The relationship between the resulting policy gradient update and the advantage calculation is defined as:
\begin{equation}
\small
\begin{aligned}
\nabla \theta &\approx \frac{1}{G} \sum_{i=1}^{G} \hat{A}_{i,g} \nabla \log \pi_\theta(a_i|s) \\
\hat{A}_{i,g} &= \frac{R_i - \text{mean}(\{R_1, \dots, R_G\})}{\text{std}(\{R_1, \dots, R_G\})}
\end{aligned}
\end{equation}
When using a standard $[0, 1]$ float reward, if all samples in a group are sub-optimal (e.g., scores $S \in [0.7, 0.9]$), a sample with a score of $0.9$—despite containing logical flaws—will still receive a positive advantage signal because it exceeds the group mean. This leads to the Non-negative reward trap, where the model tends to converge to a sub-optimal policy, learning to produce ambiguous "nonsense" that contains partially correct keywords while losing the motivation to explore perfect logic chains.

\paragraph{Reward Cliff and the Bipolar Mapping Logic.}
To correct this bias, BFR sets a strict standard: only a completely correct answer receives the sole positive reward of $1.0$. For any imperfect answer, we apply a transformation by subtracting 1 from its initial correctness score $S$ (where $S \in [0, 1)$), mapping it into the penalty interval $[-1, 0)$. This $S-1$ transformation creates a significant Reward Cliff between a perfect response ($1.0$) and all imperfect ones. This design ensures that even a near-perfect answer receives a minor penalty, mathematically enforcing the logic that "anything less than perfect is wrong," thereby driving the model toward the global optimum.

\paragraph{Penalty-Driven Correction (Push-Pull Dynamics).}
BFR constructs an efficient dual-directional dynamic mechanism. In the policy gradient update process, BFR provides differentiated negative gradients for incorrect reasoning paths. Unlike binary rewards, where incorrect samples often provide only zero signals with extremely low information density, the explicit negative rewards in BFR generate a stronger "push" force via the resulting advantages, which varies based on the degree of deviation. This "Push (negative penalty) and Pull (positive reward)" combination maximizes the training efficiency of every sampled instance, guiding the model to correct subtle logical loopholes and efficiently approximate the global optimum of logical reasoning.

\section{Experimental Setup}

To evaluate the \textsc{UltraLogic} framework and BFR mechanism, we conduct ablation studies using Qwen3-8B and 14B models.

\subsection{Training Configuration and Benchmarks}

\paragraph{Training.} All training utilizes the GRPO algorithm with consistent hyperparameters: $lr=1e-6$, rollout $=16$, max\_response\_length $=32,768$, $temperature=1.0$, and $top\_p=1.0$. The rewards in the experiments include a 0.1 format bonus to isolate logical correctness from output formatting. All reported experimental results in~\Cref{Results and Analysis} were obtained after training for two epochs.

\paragraph{Evaluation.} We evaluate models on five benchmarks: AIME (2024 \& 2025)\footnote{\url{https://artofproblemsolving.com/wiki/index.php/AIME_Problems_and_Solutions}}, HMMT 2025\footnote{\url{https://www.hmmt.org/www/archive/problems}}, BBH~\cite{suzgun2022challenging}, BBEH~\cite{kazemi-etal-2025-big}, and ARC-AGI\footnote{\url{https://arcprize.org/arc-agi/1/}}. Evaluation employs a sampling strategy ($T=0.6$, \texttt{top\_p}=0.95, \texttt{top\_k}=20, \texttt{min\_p}=0) with a 32,768 token limit ; the primary metric is accuracy averaged over 64 samples[cite: 4510].

\subsection{Ablation Study Design}
\paragraph{Difficulty Matching.} This study investigates the correlation between model capacity and optimal training difficulty. We train 8B and 14B models on three difficulty-stratified datasets (50 tasks, 10,000 samples each): Easy (levels 1--4), Medium (4--7), and Hard (7--10). Models in this study are trained using a standard $\{0, 1\}$ binary reward.

\paragraph{Bipolar Float Reward.} To validate BFR, we compare three reward schemes: Binary ($\{0, 1\}$), Graded Float ($[0, 1]$), and BFR ($[-1, 0) \cup \{1\}$). Each experiment uses 50 tasks (10k samples), with Qwen3-8B trained on Easy-only data. 

\section{Results and Analysis}

\label{Results and Analysis}
In this chapter, we provide an in-depth analysis of the experimental performance of \textsc{UltraLogic} across different dimensions. Based on the experimental setup described previously, we conducted the Difficulty Matching ablation study and the Bipolar Float Reward ablation study. We systematically reveal the impacts of difficulty matching mechanisms, and the Bipolar Float Reward on model reasoning capabilities. The details of results are shown in~\Cref{sec:appendix_exp_details}

\subsection{Difficulty Matching: The Scaling Law of Difficulty}

We identified the Difficulty Matching Phenomenon in the Qwen3-8B and 14B: model capacity dictates sensitivity to training data difficulty, exhibiting a strong positive correlation.

\begin{table}[h]
\centering
\normalsize
\renewcommand{\arraystretch}{1.0}
\setlength{\tabcolsep}{3pt}
\resizebox{\columnwidth}{!}{
\begin{tabular}{lcccccc}
\toprule
\textbf{Model} & \textbf{AIME24} & \textbf{AIME25} & \textbf{HMMT25} & \textbf{BBH} & \textbf{BBEH} & \textbf{ARC-AGI} \\
\midrule
Qwen3-8B  & 75.2 & 66.1 & 47.1 & 88.2 & 29.2 & 4.0 \\
\textbf{Easy}      & \textbf{81.7} & \textbf{69.1} & \textbf{52.3} & \textbf{90.2} & \textbf{31.1} & \textbf{4.6} \\
\underline{Medium}    & \underline{77.6} & \underline{67.6} & \underline{50.9} & \underline{88.7} & \underline{29.6} & \underline{4.2} \\
Hard      & 76.5 & 65.1 & 50.0 & 88.3 & 29.0 & 4.0 \\
\midrule
Qwen3-14B & 76.7 & 70.3 & 54.3 & 89.6 & 32.2 & 6.3 \\
Easy      & 78.8 & 72.5 & 58.6 & \underline{90.9} & 33.7 & 6.1 \\
\textbf{Medium}    & \textbf{82.4} & \textbf{75.8} & \textbf{63.1} & \textbf{92.1} & \textbf{36.7} & \textbf{7.9} \\
\underline{Hard}      & \underline{80.8} &	\underline{73.8} &	\underline{59.9} &	\underline{90.9} &	\underline{34.7}   &	\underline{6.8}  \\
\bottomrule
\end{tabular}
}
\caption{Ablation study on model scale and task difficulty.}
\label{tab:model_ablation}
\end{table}

The experimental results present distinct distribution characteristics: for the Qwen3-8B model, the training gains follow the order of Easy > Medium > Hard; whereas for the Qwen3-14B model, the gains follow the order of Medium > Hard > Easy. Through comprehensive analysis of the average score in training process, we observed that models of different sizes generally achieve the best training results on problems where their success rate—after subtracting approximately 0.1 to account for formatting accuracy—falls within the 40\% to 60\% range. This further proves the core value of the fine-grained difficulty grading function provided by the \textsc{UltraLogic} framework in adapting to the training of models of different scales.

Furthermore, regarding the stability of the training process (detailed results and figures can be seen in~\Cref{Difficulty Matching Ablation Study}, the most compatible difficulty not only brings more significant performance improvements but also results in smoother and more prominent training curves. Conversely, overly simple problems fail to provide sufficient information increment, resulting in negligible improvements to the model. Meanwhile, high-difficulty problems that exceed the model's capability boundaries tend to introduce excessive noise and may even have an adverse effect, potentially leading to training collapse. This observation highlights the importance of reinforcement learning within the ``Zone of Proximal Development,'' where the effectiveness of gradient signals is maximized only when the task difficulty matches the model's current cognitive level. 

\subsection{Effectiveness of Bipolar Float Reward: The Penalty-Driven Optimization}

To validate the BFR mechanism's role in enhancing reasoning precision, we conducted a comparative ablation study on the Qwen3-8B model against Binary Reward ($\{0, 1\}$) and standard Graded Float Reward ($[0, 1]$). 

\begin{table}[h]
\centering
\normalsize
\renewcommand{\arraystretch}{1.0}
\setlength{\tabcolsep}{3pt}
\resizebox{\columnwidth}{!}{
\begin{tabular}{lcccccc}
\toprule
\textbf{Reward} & \textbf{AIME24} & \textbf{AIME25} & \textbf{HMMT25} & \textbf{BBH} & \textbf{BBEH} & \textbf{ARC-AGI} \\
\midrule
Qwen3-8B & 75.2 & 66.1 & 47.1 & 88.2 & 29.2 & 4.0 \\
Binary Reward & \underline{81.7} & \underline{69.1} & 52.3 & 90.2 & \underline{31.1} & \underline{4.6} \\
Graded Float & 76.9 &	66.3 &	\underline{53.0} &	\underline{90.4} &	31.0 &	4.3 \\
\textbf{Bipolar Float} & \textbf{82.6} &	\textbf{71.3} &	\textbf{56.6} &	\textbf{91.1} &	\textbf{32.5} &	\textbf{4.7} \\

\bottomrule
\end{tabular}
}
\caption{Ablation study on reward mechanisms.}
\label{tab:reward_ablation}
\end{table}

\vspace{-10pt}

\begin{figure}[htbp]
    \centering
    \includegraphics[width=0.85\linewidth]{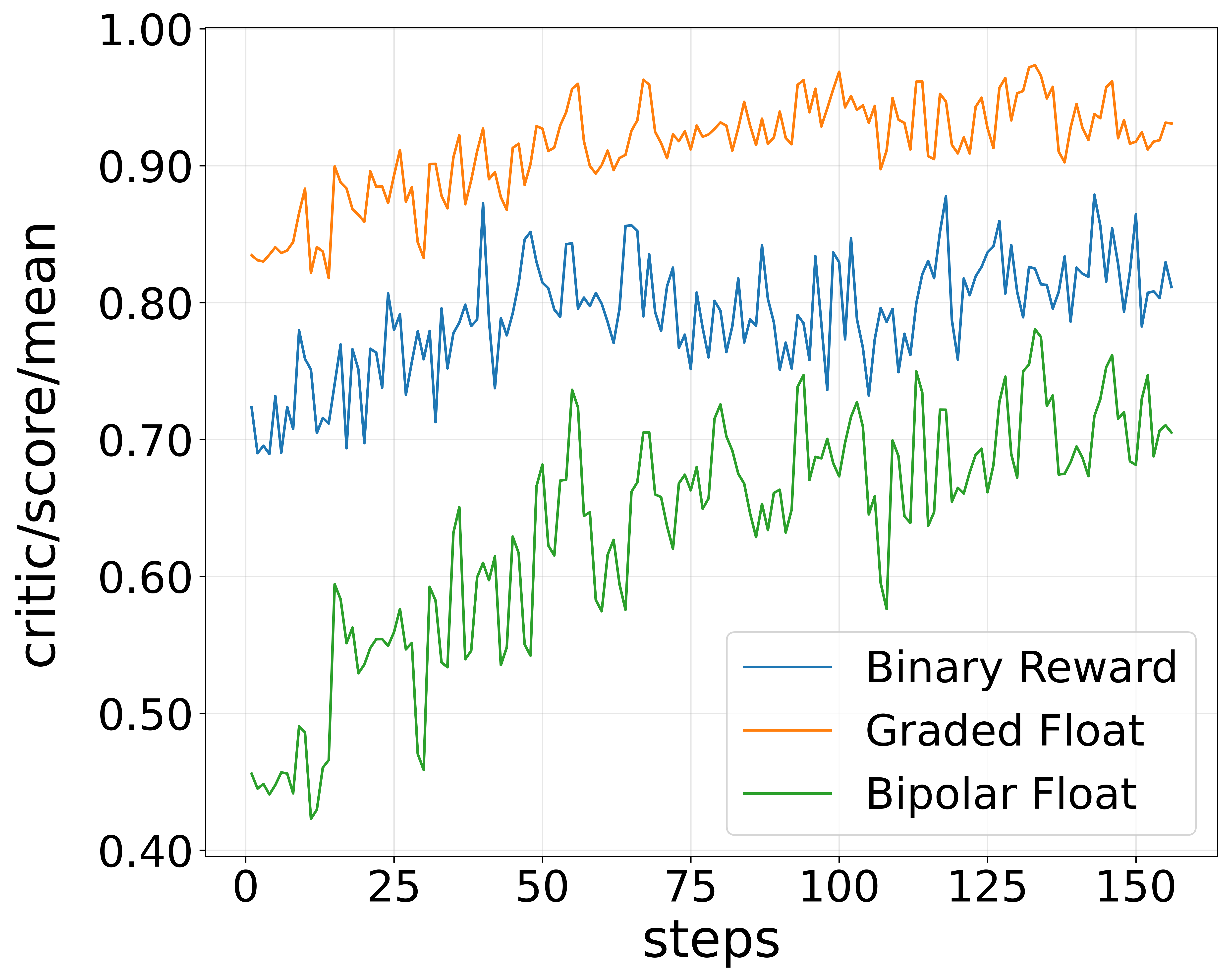}
    \caption{The critic/score/mean metrics of Qwen3-8B during the GRPO process by using different reward mechanisms.}
    \label{fig:reward_metrics}
\end{figure}

As shown in~\Cref{tab:reward_ablation}, the model trained with BFR achieved optimal performance across nearly all benchmarks. BFR significantly outperformed the Binary baseline and surpassed Graded Float in both convergence speed and final accuracy. Particularly in logic-intensive tasks like AIME and BBH, BFR's gains confirm that graded penalties are critical for breaking through performance bottlenecks in complex reasoning tasks. The dynamic progression of training metrics, illustrated in~\Cref{fig:reward_metrics}, further demonstrates BFR's superior efficiency in providing clearer optimization signals to the policy network. Detailed training convergence curves and expanded performance tables can be found in~\Cref{Bipolar Float Reward Ablation Study}.

This substantial improvement is attributed to BFR's ability to resolve inherent defects in traditional rewards. As discussed in~\Cref{sec:bfr_logic}, BFR breaks the Non-negative reward trap and implements a ``Push-Pull'' optimization mechanism, preventing convergence to sub-optimal policies and guiding the model efficiently toward the global logical optimum.

\subsection{Additional Empirical Observations}

Beyond the primary ablation studies, we identify two key empirical findings regarding the stability of the Reinforcement Learning (RL) process:

\paragraph{Data Quality Sensitivity.} Compared to Supervised Fine-Tuning (SFT), RLVR is remarkably intolerant of noise. In our experiments, if even 1--3 task types out of 50 contain logic errors in the solution or template, the model invariably suffers from training collapse. This ``brittle quality threshold'' confirms that the validation gate described in~\Cref{Data Validation and Quality Assurance} is an absolute prerequisite for successful training.

\paragraph{Architectural Choice.} Initial trials with Mixture-of-Experts (MoE) architectures showed frequent divergence during the GRPO process. To ensure stable gradient dynamics and smooth convergence, we transitioned to Dense models (e.g., Qwen3-8B and 14B), which proved significantly more robust in handling the complex reasoning signals produced by the \textsc{UltraLogic} framework.

\section{Conclusion}

We present the \textsc{UltraLogic} data framework and the Bipolar Float Reward (BFR) mechanism. \textsc{UltraLogic} employs an automated pipeline—combining seed tasks with programmatic expansion to enable the large-scale synthesis of diverse, verifiable, and difficulty-stratified training datasets. Our experiments identify a difficulty matching phenomenon: training efficiency is maximized when task difficulty aligns with the model's zone of proximal development. Furthermore, the BFR mechanism addresses the inherent information sparsity and the Non-negative reward trap found in traditional reward structures. This approach enhances the precision of reward signals, thereby improving both training accuracy and optimization efficiency.

\section*{Limitations}
\paragraph{Dependence on Human Annotation.}
Despite the high degree of automation achieved by \textsc{UltraLogic}, the extreme precision required for logical reasoning tasks necessitates manual intervention from human annotators in key stages, such as seed task logic verification and initial difficulty calibration. Since reinforcement learning with verifiable rewards is remarkably intolerant of noise, even minor logical flaws in the synthetic data can lead to training collapse. Consequently, while we strive for full automation, this reliance on annotators remains a necessary trade-off to ensure 100\% logical rigor and to maintain the "gold standard" quality of the reasoning dataset.

\paragraph{Heuristic Nature of Reward Scaling.} 
Although the Bipolar Float Reward (BFR) mechanism successfully guides the model toward global logical optima through advantage reshaping, the current configuration of reward values remains largely based on intuitive heuristics. Theoretically, for tasks with varying logical depths, there should exist more precise and fine-grained non-integer reward signals that could provide superior guidance. However, in the absence of a universal methodology to automatically search for "mathematically perfect" reward values, we have adopted these robust and intuitive settings to ensure the stability of gradient dynamics across diverse reasoning scenarios.


\bibliography{main}

@misc{liu2025synlogicsynthesizingverifiablereasoning,
      title={SynLogic: Synthesizing Verifiable Reasoning Data at Scale for Learning Logical Reasoning and Beyond}, 
      author={Junteng Liu and Yuanxiang Fan and Zhuo Jiang and Han Ding and Yongyi Hu and Chi Zhang and Yiqi Shi and Shitong Weng and Aili Chen and Shiqi Chen and Yunan Huang and Mozhi Zhang and Pengyu Zhao and Junjie Yan and Junxian He},
      year={2025},
      eprint={2505.19641},
      archivePrefix={arXiv},
      primaryClass={cs.AI},
      url={https://arxiv.org/abs/2505.19641}, 
}

@misc{liu2025agenticmathenhancingllmreasoning,
      title={AgenticMath: Enhancing LLM Reasoning via Agentic-based Math Data Generation}, 
      author={Xianyang Liu and Yilin Liu and Shuai Wang and Hao Cheng and Andrew Estornell and Yuzhi Zhao and Jiaheng Wei},
      year={2025},
      eprint={2510.19361},
      archivePrefix={arXiv},
      primaryClass={cs.CL},
      url={https://arxiv.org/abs/2510.19361}, 
}

@article{yu2024metamathbootstrapmathematicalquestions,
  title={MetaMath: Bootstrap Your Own Mathematical Questions for Large Language Models},
  author={Yu, Longhui and Jiang, Weisen and Shi, Han and Yu, Jincheng and Liu, Zhengying and Zhang, Yu and Kwok, James T and Li, Zhenguo and Weller, Adrian and Liu, Weiyang},
  journal={arXiv preprint arXiv:2309.12284},
  year={2023}
}

@inproceedings{lu2024mathgeniegeneratingsyntheticdata,
    title = "{M}ath{G}enie: Generating Synthetic Data with Question Back-translation for Enhancing Mathematical Reasoning of {LLM}s",
    author = "Lu, Zimu  and
      Zhou, Aojun  and
      Ren, Houxing  and
      Wang, Ke  and
      Shi, Weikang  and
      Pan, Junting  and
      Zhan, Mingjie  and
      Li, Hongsheng",
    editor = "Ku, Lun-Wei  and
      Martins, Andre  and
      Srikumar, Vivek",
    booktitle = "Proceedings of the 62nd Annual Meeting of the Association for Computational Linguistics (Volume 1: Long Papers)",
    month = aug,
    year = "2024",
    address = "Bangkok, Thailand",
    publisher = "Association for Computational Linguistics",
    url = "https://aclanthology.org/2024.acl-long.151/",
    doi = "10.18653/v1/2024.acl-long.151",
    pages = "2732--2747"
}

@inproceedings{
lightman2023letsverifystepstep,
title={Let's Verify Step by Step},
author={Hunter Lightman and Vineet Kosaraju and Yuri Burda and Harrison Edwards and Bowen Baker and Teddy Lee and Jan Leike and John Schulman and Ilya Sutskever and Karl Cobbe},
booktitle={The Twelfth International Conference on Learning Representations},
year={2024},
url={https://openreview.net/forum?id=v8L0pN6EOi}
}

@misc{ma2023letsrewardstepstep,
      title={Let's reward step by step: Step-Level reward model as the Navigators for Reasoning}, 
      author={Qianli Ma and Haotian Zhou and Tingkai Liu and Jianbo Yuan and Pengfei Liu and Yang You and Hongxia Yang},
      year={2023},
      eprint={2310.10080},
      archivePrefix={arXiv},
      primaryClass={cs.CL},
      url={https://arxiv.org/abs/2310.10080}, 
}

@inproceedings{
zhang2025openprm,
title={Open{PRM}: Building Open-domain Process-based Reward Models with Preference Trees},
author={Kaiyan Zhang and Jiayuan Zhang and Haoxin Li and Xuekai Zhu and Ermo Hua and Xingtai Lv and Ning Ding and Biqing Qi and Bowen Zhou},
booktitle={The Thirteenth International Conference on Learning Representations},
year={2025},
url={https://openreview.net/forum?id=fGIqGfmgkW}
}

@inproceedings{yin2025dynamicgeneralizableprocessreward,
  title={Dynamic and generalizable process reward modeling},
  author={Yin, Zhangyue and Sun, Qiushi and Zeng, Zhiyuan and Cheng, Qinyuan and Qiu, Xipeng and Huang, Xuan-Jing},
  booktitle={Proceedings of the 63rd Annual Meeting of the Association for Computational Linguistics (Volume 1: Long Papers)},
  pages={4203--4233},
  year={2025}
}

@article{kwan2025opensir,
  title={OpenSIR: Open-Ended Self-Improving Reasoner},
  author={Kwan, Wai-Chung and Leang, Joshua Ong Jun and Vougiouklis, Pavlos and Pan, Jeff Z and Valentino, Marco and Minervini, Pasquale},
  journal={arXiv preprint arXiv:2511.00602},
  year={2025}
}

@article{wang2025morphobench,
  title={MorphoBench: A Benchmark with Difficulty Adaptive to Model Reasoning},
  author={Wang, Xukai and Liu, Xuanbo and Chen, Mingrui and Zhong, Haitian and Yang, Xuanlin and Zeng, Bohan and Hu, Jinbo and Liang, Hao and Niu, Junbo and Li, Xuchen and others},
  journal={arXiv preprint arXiv:2510.14265},
  year={2025}
}

@inproceedings{liao2025enhancingefficiencyexplorationreinforcement,
    title = "Enhancing Efficiency and Exploration in Reinforcement Learning for {LLM}s",
    author = "Liao, Mengqi  and
      Xi, Xiangyu  and
      Ruinian, Chen  and
      Leng, Jia  and
      Hu, Yangen  and
      Zeng, Ke  and
      Liu, Shuai  and
      Wan, Huaiyu",
    editor = "Christodoulopoulos, Christos  and
      Chakraborty, Tanmoy  and
      Rose, Carolyn  and
      Peng, Violet",
    booktitle = "Proceedings of the 2025 Conference on Empirical Methods in Natural Language Processing",
    month = nov,
    year = "2025",
    address = "Suzhou, China",
    publisher = "Association for Computational Linguistics",
    url = "https://aclanthology.org/2025.emnlp-main.75/",
    doi = "10.18653/v1/2025.emnlp-main.75",
    pages = "1451--1463",
    ISBN = "979-8-89176-332-6"
}

@misc{li2025stayingsweetspotresponsive,
      title={Staying in the Sweet Spot: Responsive Reasoning Evolution via Capability-Adaptive Hint Scaffolding}, 
      author={Ziheng Li and Zexu Sun and Jinman Zhao and Erxue Min and Yongcheng Zeng and Hui Wu and Hengyi Cai and Shuaiqiang Wang and Dawei Yin and Xu Chen and Zhi-Hong Deng},
      year={2025},
      eprint={2509.06923},
      archivePrefix={arXiv},
      primaryClass={cs.LG},
      url={https://arxiv.org/abs/2509.06923}, 
}

@article{achiam2023gpt,
  title={Gpt-4 technical report},
  author={Achiam, Josh and Adler, Steven and Agarwal, Sandhini and Ahmad, Lama and Akkaya, Ilge and Aleman, Florencia Leoni and Almeida, Diogo and Altenschmidt, Janko and Altman, Sam and Anadkat, Shyamal and others},
  journal={arXiv preprint arXiv:2303.08774},
  year={2023}
}

@misc{deepseekai2025deepseekv3technicalreport,
      title={DeepSeek-V3 Technical Report}, 
      author={DeepSeek-AI and Aixin Liu and Bei Feng and Bing Xue and Bingxuan Wang and Bochao Wu and Chengda Lu and Chenggang Zhao and Chengqi Deng and Chenyu Zhang and Chong Ruan and Damai Dai and Daya Guo and Dejian Yang and Deli Chen and Dongjie Ji and Erhang Li and Fangyun Lin and Fucong Dai and Fuli Luo and Guangbo Hao and Guanting Chen and Guowei Li and H. Zhang and Han Bao and Hanwei Xu and Haocheng Wang and Haowei Zhang and Honghui Ding and Huajian Xin and Huazuo Gao and Hui Li and Hui Qu and J. L. Cai and Jian Liang and Jianzhong Guo and Jiaqi Ni and Jiashi Li and Jiawei Wang and Jin Chen and Jingchang Chen and Jingyang Yuan and Junjie Qiu and Junlong Li and Junxiao Song and Kai Dong and Kai Hu and Kaige Gao and Kang Guan and Kexin Huang and Kuai Yu and Lean Wang and Lecong Zhang and Lei Xu and Leyi Xia and Liang Zhao and Litong Wang and Liyue Zhang and Meng Li and Miaojun Wang and Mingchuan Zhang and Minghua Zhang and Minghui Tang and Mingming Li and Ning Tian and Panpan Huang and Peiyi Wang and Peng Zhang and Qiancheng Wang and Qihao Zhu and Qinyu Chen and Qiushi Du and R. J. Chen and R. L. Jin and Ruiqi Ge and Ruisong Zhang and Ruizhe Pan and Runji Wang and Runxin Xu and Ruoyu Zhang and Ruyi Chen and S. S. Li and Shanghao Lu and Shangyan Zhou and Shanhuang Chen and Shaoqing Wu and Shengfeng Ye and Shengfeng Ye and Shirong Ma and Shiyu Wang and Shuang Zhou and Shuiping Yu and Shunfeng Zhou and Shuting Pan and T. Wang and Tao Yun and Tian Pei and Tianyu Sun and W. L. Xiao and Wangding Zeng and Wanjia Zhao and Wei An and Wen Liu and Wenfeng Liang and Wenjun Gao and Wenqin Yu and Wentao Zhang and X. Q. Li and Xiangyue Jin and Xianzu Wang and Xiao Bi and Xiaodong Liu and Xiaohan Wang and Xiaojin Shen and Xiaokang Chen and Xiaokang Zhang and Xiaosha Chen and Xiaotao Nie and Xiaowen Sun and Xiaoxiang Wang and Xin Cheng and Xin Liu and Xin Xie and Xingchao Liu and Xingkai Yu and Xinnan Song and Xinxia Shan and Xinyi Zhou and Xinyu Yang and Xinyuan Li and Xuecheng Su and Xuheng Lin and Y. K. Li and Y. Q. Wang and Y. X. Wei and Y. X. Zhu and Yang Zhang and Yanhong Xu and Yanhong Xu and Yanping Huang and Yao Li and Yao Zhao and Yaofeng Sun and Yaohui Li and Yaohui Wang and Yi Yu and Yi Zheng and Yichao Zhang and Yifan Shi and Yiliang Xiong and Ying He and Ying Tang and Yishi Piao and Yisong Wang and Yixuan Tan and Yiyang Ma and Yiyuan Liu and Yongqiang Guo and Yu Wu and Yuan Ou and Yuchen Zhu and Yuduan Wang and Yue Gong and Yuheng Zou and Yujia He and Yukun Zha and Yunfan Xiong and Yunxian Ma and Yuting Yan and Yuxiang Luo and Yuxiang You and Yuxuan Liu and Yuyang Zhou and Z. F. Wu and Z. Z. Ren and Zehui Ren and Zhangli Sha and Zhe Fu and Zhean Xu and Zhen Huang and Zhen Zhang and Zhenda Xie and Zhengyan Zhang and Zhewen Hao and Zhibin Gou and Zhicheng Ma and Zhigang Yan and Zhihong Shao and Zhipeng Xu and Zhiyu Wu and Zhongyu Zhang and Zhuoshu Li and Zihui Gu and Zijia Zhu and Zijun Liu and Zilin Li and Ziwei Xie and Ziyang Song and Ziyi Gao and Zizheng Pan},
      year={2025},
      eprint={2412.19437},
      archivePrefix={arXiv},
      primaryClass={cs.CL},
      url={https://arxiv.org/abs/2412.19437}, 
}

@misc{yang2025qwen3technicalreport,
      title={Qwen3 Technical Report}, 
      author={An Yang and Anfeng Li and Baosong Yang and Beichen Zhang and Binyuan Hui and Bo Zheng and Bowen Yu and Chang Gao and Chengen Huang and Chenxu Lv and Chujie Zheng and Dayiheng Liu and Fan Zhou and Fei Huang and Feng Hu and Hao Ge and Haoran Wei and Huan Lin and Jialong Tang and Jian Yang and Jianhong Tu and Jianwei Zhang and Jianxin Yang and Jiaxi Yang and Jing Zhou and Jingren Zhou and Junyang Lin and Kai Dang and Keqin Bao and Kexin Yang and Le Yu and Lianghao Deng and Mei Li and Mingfeng Xue and Mingze Li and Pei Zhang and Peng Wang and Qin Zhu and Rui Men and Ruize Gao and Shixuan Liu and Shuang Luo and Tianhao Li and Tianyi Tang and Wenbiao Yin and Xingzhang Ren and Xinyu Wang and Xinyu Zhang and Xuancheng Ren and Yang Fan and Yang Su and Yichang Zhang and Yinger Zhang and Yu Wan and Yuqiong Liu and Zekun Wang and Zeyu Cui and Zhenru Zhang and Zhipeng Zhou and Zihan Qiu},
      year={2025},
      eprint={2505.09388},
      archivePrefix={arXiv},
      primaryClass={cs.CL},
      url={https://arxiv.org/abs/2505.09388}, 
}

@inproceedings{wu-etal-2025-c2rbench,
    title = "{C}{\texttwosuperior}{RB}ench: A {C}hinese Complex Reasoning Benchmark for Large Language Models",
    author = "Wu, Junru  and
      Shen, Tianhao  and
      Su, Linxi  and
      Xiong, Deyi",
    editor = "Che, Wanxiang  and
      Nabende, Joyce  and
      Shutova, Ekaterina  and
      Pilehvar, Mohammad Taher",
    booktitle = "Findings of the Association for Computational Linguistics: ACL 2025",
    month = jul,
    year = "2025",
    address = "Vienna, Austria",
    publisher = "Association for Computational Linguistics",
    url = "https://aclanthology.org/2025.findings-acl.1083/",
    doi = "10.18653/v1/2025.findings-acl.1083",
    pages = "21031--21050",
    ISBN = "979-8-89176-256-5"
}

@article{ouyang2022traininglanguagemodelsfollow,
  title={Training language models to follow instructions with human feedback},
  author={Ouyang, Long and Wu, Jeffrey and Jiang, Xu and Almeida, Diogo and Wainwright, Carroll and Mishkin, Pamela and Zhang, Chong and Agarwal, Sandhini and Slama, Katarina and Ray, Alex and others},
  journal={Advances in neural information processing systems},
  volume={35},
  pages={27730--27744},
  year={2022}
}

@article{rafailov2024directpreferenceoptimizationlanguage,
  title={Direct preference optimization: Your language model is secretly a reward model},
  author={Rafailov, Rafael and Sharma, Archit and Mitchell, Eric and Manning, Christopher D and Ermon, Stefano and Finn, Chelsea},
  journal={Advances in neural information processing systems},
  volume={36},
  pages={53728--53741},
  year={2023}
}

@misc{shao2024deepseekmathpushinglimitsmathematical,
      title={DeepSeekMath: Pushing the Limits of Mathematical Reasoning in Open Language Models}, 
      author={Zhihong Shao and Peiyi Wang and Qihao Zhu and Runxin Xu and Junxiao Song and Xiao Bi and Haowei Zhang and Mingchuan Zhang and Y. K. Li and Y. Wu and Daya Guo},
      year={2024},
      eprint={2402.03300},
      archivePrefix={arXiv},
      primaryClass={cs.CL},
      url={https://arxiv.org/abs/2402.03300}, 
}

@article{hendrycks2021measuringmathematicalproblemsolving,
  title={Measuring Mathematical Problem Solving With the MATH Dataset},
  author={Dan Hendrycks and Collin Burns and Saurav Kadavath and Akul Arora and Steven Basart and Eric Tang and Dawn Song and Jacob Steinhardt},
  journal={NeurIPS},
  year={2021}
}

@inproceedings{
rein2023gpqagraduatelevelgoogleproofqa,
title={{GPQA}: A Graduate-Level Google-Proof Q\&A Benchmark},
author={David Rein and Betty Li Hou and Asa Cooper Stickland and Jackson Petty and Richard Yuanzhe Pang and Julien Dirani and Julian Michael and Samuel R. Bowman},
booktitle={First Conference on Language Modeling},
year={2024},
url={https://openreview.net/forum?id=Ti67584b98}
}

@misc{li202512surveyreasoning,
      title={From System 1 to System 2: A Survey of Reasoning Large Language Models}, 
      author={Zhong-Zhi Li and Duzhen Zhang and Ming-Liang Zhang and Jiaxin Zhang and Zengyan Liu and Yuxuan Yao and Haotian Xu and Junhao Zheng and Pei-Jie Wang and Xiuyi Chen and Yingying Zhang and Fei Yin and Jiahua Dong and Zhiwei Li and Bao-Long Bi and Ling-Rui Mei and Junfeng Fang and Xiao Liang and Zhijiang Guo and Le Song and Cheng-Lin Liu},
      year={2025},
      eprint={2502.17419},
      archivePrefix={arXiv},
      primaryClass={cs.AI},
      url={https://arxiv.org/abs/2502.17419}, 
}

@inproceedings{mishra2023lilaunifiedbenchmarkmathematical,
    title = "{LILA}: A Unified Benchmark for Mathematical Reasoning",
    author = "Mishra, Swaroop  and
      Finlayson, Matthew  and
      Lu, Pan  and
      Tang, Leonard  and
      Welleck, Sean  and
      Baral, Chitta  and
      Rajpurohit, Tanmay  and
      Tafjord, Oyvind  and
      Sabharwal, Ashish  and
      Clark, Peter  and
      Kalyan, Ashwin",
    editor = "Goldberg, Yoav  and
      Kozareva, Zornitsa  and
      Zhang, Yue",
    booktitle = "Proceedings of the 2022 Conference on Empirical Methods in Natural Language Processing",
    month = dec,
    year = "2022",
    address = "Abu Dhabi, United Arab Emirates",
    publisher = "Association for Computational Linguistics",
    url = "https://aclanthology.org/2022.emnlp-main.392/",
    doi = "10.18653/v1/2022.emnlp-main.392",
    pages = "5807--5832"
}

@inproceedings{
paulus2017deepreinforcedmodelabstractive,
title={A Deep Reinforced Model for Abstractive Summarization},
author={Romain Paulus and Caiming Xiong and Richard Socher},
booktitle={International Conference on Learning Representations},
year={2018},
url={https://openreview.net/forum?id=HkAClQgA-},
}

@article{yao2022react,
  title={ReAct: Synergizing Reasoning and Acting in Language Models},
  author={Yao, Shunyu and Zhao, Jeffrey and Yu, Dian and Du, Nan and Shafran, Izhak and Narasimhan, Karthik and Cao, Yuan},
  journal={arXiv preprint arXiv:2210.03629},
  year={2022}
}

@article{suzgun2022challenging,
  title={Challenging BIG-Bench Tasks and Whether Chain-of-Thought Can Solve Them},
  author={Suzgun, Mirac and Scales, Nathan and Sch{\"a}rli, Nathanael and Gehrmann, Sebastian and Tay, Yi and Chung, Hyung Won and Chowdhery, Aakanksha and Le, Quoc V and Chi, Ed H and Zhou, Denny and and Wei, Jason},
  journal={arXiv preprint arXiv:2210.09261},
  year={2022}
}

@inproceedings{kazemi-etal-2025-big,
    title = "{BIG}-Bench Extra Hard",
    author = "Kazemi, Mehran  and
      Fatemi, Bahare  and
      Bansal, Hritik  and
      Palowitch, John  and
      Anastasiou, Chrysovalantis  and
      Mehta, Sanket Vaibhav  and
      Jain, Lalit K  and
      Aglietti, Virginia  and
      Jindal, Disha  and
      Chen, Peter  and
      Dikkala, Nishanth  and
      Tyen, Gladys  and
      Liu, Xin  and
      Shalit, Uri  and
      Chiappa, Silvia  and
      Olszewska, Kate  and
      Tay, Yi  and
      Tran, Vinh Q.  and
      Le, Quoc V  and
      Firat, Orhan",
    editor = "Che, Wanxiang  and
      Nabende, Joyce  and
      Shutova, Ekaterina  and
      Pilehvar, Mohammad Taher",
    booktitle = "Proceedings of the 63rd Annual Meeting of the Association for Computational Linguistics (Volume 1: Long Papers)",
    month = jul,
    year = "2025",
    address = "Vienna, Austria",
    publisher = "Association for Computational Linguistics",
    url = "https://aclanthology.org/2025.acl-long.1285/",
    doi = "10.18653/v1/2025.acl-long.1285",
    pages = "26473--26501",
    ISBN = "979-8-89176-251-0"
}

\appendix
\section{Detailed Data Settings}
\label{sec:appendix_data_details}
This appendix provides comprehensive details regarding the \textsc{UltraLogic} dataset configuration.

\subsection{Three-dimensional Orthogonal Categories}
\label{sec:Three-dimensional Orthogonal Categories}
We employ a three-dimensional orthogonal classification system to guide task creation and screening, ensuring that the repository covers a broad spectrum of reasoning challenges. Each reasoning task can be uniquely described and situated along these three dimensions.




The complete taxonomy, including all sub-categories and their respective descriptions within the \textsc{UltraLogic} Task Repository, is presented in~\Cref{tab:taxonomy}.

\onecolumn

\begin{table}[ht]
\centering

\footnotesize

\setlength{\tabcolsep}{6pt}
\renewcommand{\arraystretch}{1.4}
  \resizebox{\linewidth}{!}{%

  \rowcolors{3}{gray!10}{white}

\begin{tabularx}{\columnwidth}{l l >{\raggedright\arraybackslash}X}
      \toprule
      \textbf{Categories} & \textbf{Sub-categories} & \textbf{Description} \\
      \midrule
      \textbf{Task Domain} & Symbolic Manipulation & Manipulate letters and special symbols, such as symbol or regular matching. \\
       & Numerical Manipulation & Manipulate numbers, such as numerical conversion or numerical decryption. \\
       & Textual Manipulation & Manipulate word or text, such as password puzzles. \\
       & Object Manipulation (Real-World) & Tasks integrated with real-world scenarios, such as case investigation. \\
       & Planning \& Scheduling & Tasks involving the planning and organization of complex tasks, such as elevator scheduling, cargo transportation. \\
       & Classic Games & Classic games such as Sudoku, 24-point, or Monopoly. \\
       & Spatial: Geometry & Describe complex geometric or positional relationships using plain text. \\
       & Spatial: Pathfinding & Pathfinding problem, finding the optimal path. \\
       & Spatial: Visual-to-Text & Numbers, letters or symbols are used to represent graphical or geometric positional relationships, such as a  two-dimensional map. \\
       & Cross-Topic & Various combinations of the aforementioned tasks. \\
       & Others & - \\
   \midrule
      \textbf{Core Ability} & Constraint Satisfaction & The problem will contain numerous conditions and constraint rules. It is essential to pay attention to ensuring that solutions do not contradict the conditions or violate the rules \\
       & Algorithmic Thinking & Solution requires the application of algorithmic concepts such as dynamic programming, the knapsack problem or depth-first search. \\
       & Info Extraction \& Integration & The problem may contain a significant amount of irrelevant information, which requires identifying the effective information relevant to solving the problem. \\
       & Item Connection \& Mapping & Identify the mapping relationships between multiple items, such as the relationships between spatial connections and cipher mappings \\
       & Instruction Following & The problem may require performing multiple operations on an item, such as rotating a Rubik's cube multiple times. When solving the problem, it is essential to follow to the specified sequence of operations. \\
       & Others & - \\
       \midrule
      \textbf{Difficulty Source} & Complex Rules & The difficulty of the problem lies in the numerous and complex rules involved, such as in complex board games. \\
       & Complex Conditions & The problem presents multiple conditional pathways for selection, requiring precise comprehension of the conditions to enable correct choices at each step and ultimately solve the problem, as exemplified in game-level progression scenarios \\
       & Large Search Space & When solving the problem requires identifying the optimal solution, it is necessary to explore numerous approaches to achieve this, such as the maze problem. \\
       & Tedious Solution Steps & The problem may contain lengthy procedural steps or complex logical structures, requiring tedious solution steps to resolve. \\
       & Computational Complexity & Problem-solving involves complex mathematical computations \\
       & Intrinsic LLM Weaknesses & Some weaknesses inherent to large models, such as errors in numeric and alphabetic characters caused by token encoding issues \\
       & Others & - \\
      \bottomrule
    \end{tabularx}
  }
\caption{The three-dimensional orthogonal classification system of the \textsc{UltraLogic} Task Repository. Each task type is synthesized by the intersection of sub-categories selected from three independent dimensions.}
\label{tab:taxonomy}
\end{table}
\clearpage

\subsection{Task Category Cases}
\label{Task Category Cases}

\begin{table}[!ht]
\centering

\footnotesize
\setlength{\tabcolsep}{5pt}
\renewcommand{\arraystretch}{1.0}
  \resizebox{\linewidth}{!}{%

  \rowcolors{2}{gray!10}{white}

\begin{tabularx}{\columnwidth}{>{\raggedright}m{0.7\textwidth}  m{0.26\textwidth}}
      \toprule

        \textbf{Task examples} & \makecell[l]{\textbf{Task Domain} \\ \textbf{Core Ability} \\ \textbf{Difficulty Source}} \\ 

\midrule
Observe the chat records of a certain study group. There are 7 members who sent messages one after another, and the content is: Wright: There are exactly 6 people telling the truth. Turner: There are at least 6 people telling the lie. Ross: There are at least 4 people telling the lie. Torres: There are at least 3 people telling the lie. Harris: There are exactly 3 people telling the lie. Brooks: There are at least 2 people telling the lie. Garcia: There are at least 1 people telling the truth. Question: Among these 7 members who sent messages, who are the ones telling the truth?  & \makecell[l]{Object Manipulation (Real-World)\\ Constraint Satisfaction\\ Large Search Space}\\ 
\midrule

In an ancient and mysterious maze, an adventurer wants to find the exit. The maze is a two-dimensional layout, described as follows: \\ 
$\square\square\square\blacksquare\square$ \\
$\blacksquare\square\square\blacksquare\blacksquare$ \\
$\blacksquare\blacksquare\square\blacksquare\blacksquare$ \\
$\blacksquare\blacksquare\square\square\blacksquare$ \\
$\blacksquare\square\blacksquare\square\square$ \\
The entrance of the maze is in the top-left corner, and the exit is in the bottom-right corner. There are some exploration paths, where each exploration action represents moving 1 unit in the specified direction. If the corresponding direction is blocked by a maze wall, this step is skipped. The paths are as follows:  \\
A:right, down, right, down, right, down, right \\
B:right, down, right, down, down, right, down, down, right \\
C:right, down, down, right, down, right \\
D:right, down, down, right, down, down, down, right, down, right \\
Please list all the path IDs that allow the adventurer to get from the start point to the endpoint and exit the maze & \makecell[l]{Spatial: Pathfinding\\Instruction Following\\ Tedious Solution Steps} \\ 

\midrule
The Royal Antiquities Museum has enlisted a legendary cipher expert to reassign symbolic seals for its treasured artifacts. The original seals consist of interleaved uppercase letters, lowercase letters, and positive/negative integers. \\
 The specific seal to be decoded is: 6889SnKBUNXl. \\
 Decoding rules are as follows:   \\
1. Eliminate all lowercase letters. \\
  2. Retain uppercase letters in their original sequence.   \\
3. Sum all integers (defined as consecutive digits) to form the numerical suffix of the new seal.   \\
What is the length of the newly assigned seal after applying these rules?   & \makecell[l]{Symbolic Manipulation\\Instruction Following\\Intrinsic LLM Weaknesses} \\ 

\midrule

In the art summer camp, the student Lily is creating an abstract painting. The canvas is a grid composed of small squares, and each square must be painted with a specific color. Lily's rule is that each time she can only use one color to paint a complete rectangular area, and the same color cannot be reused. Now given the target color distribution of each position on the canvas:  \\
row 1: color: 1, 1, 1 \\
row 2: color: 1, 1, 3 \\
row 3: color: 1, 2, 1 \\
can Lily complete the painting according to the rules? & \makecell[l]{Spatial: Geometry\\Constraint Satisfaction\\Large Search Space} \\

\midrule

In a particular location, scorching weather triggered a series of cascading effects. Determine the valid causal chain based on the following conditions.   \\
**Events**:   \\
  B17: High-speed closure \\
  A1: Heavy rain \\
  C9: The driver failed to avoid the space station data \\
**Conditions**:   \\
  (1) If there is no "heavy rain", "highway closure" is unlikely to occur.\\
  (2) Typically, [highway closures] will occur shortly after [heavy rain].\\
  (3) Without "highway closure", it is unlikely for "drivers to fail to avoid" accidents. \\
  (4) Typically, after [high-speed closure], [driver's failure to avoid] will occur shortly thereafter.  \\
**Question**: What is the established causal chain? & \makecell[l]{Planning \& Scheduling \\ Info Extraction \& Integration \\ Complex Conditions} \\
\bottomrule
    \end{tabularx}
}
\caption{Representative task examples across different domains, core abilities, and difficulty sources in the \textsc{UltraLogic} framework.}
\label{tab:task_examples}
\end{table}

\clearpage

\subsection{A Complete Task Case}
\label{A Complete Task Case}
\begingroup
\setlength{\abovecaptionskip}{15pt}
\footnotesize
\setlength{\tabcolsep}{5pt}
\renewcommand{\arraystretch}{1.0}
\rowcolors{2}{gray!10}{white}

\begin{longtable}{p{0.15\textwidth} p{0.80\textwidth}}
\toprule
\rowcolors{2}{gray!10}{white}
\textbf{Field} & \textbf{Content} \\
\midrule
\endfirsthead
\toprule
\rowcolors{2}{gray!10}{white}
\textbf{Field} & \textbf{Content} (Continued) \\
\midrule
\endhead
\bottomrule
\endfoot
\bottomrule
\noalign{\vspace{10pt}}
\caption{Complete Task Case: True and False Game (Full Metadata)} \label{tab:complete_task_case} \\
\endlastfoot

\textit{Task ID} & \texttt{true\_and\_false\_game} \\
\midrule
\rowcolor{white} \textit{Problem Case} & Observe the chat records of a certain study group. There are 7 members who sent messages one after another, and the content is: Wright: There are exactly 6 people telling the truth. Turner: There are at least 6 people telling the lie. Ross: There are at least 4 people telling the lie. Torres: There are at least 3 people telling the lie. Harris: There are exactly 3 people telling the lie. Brooks: There are at least 2 people telling the lie. Garcia: There are at least 1 people telling the truth. Question: Among these 7 members who sent messages, who are the ones telling the truth? \\
\midrule
\textit{Categories} & Object Manipulation (Real-World) \& Constraint Satisfaction \& Large Search Space \\
\midrule
\rowcolor{white} \textit{En Templates} & 
\begin{enumerate}[leftmargin=*, nosep, label=\arabic*.]
    \item "At the discussion meeting of the new product project team of a certain company, the participants took turns expressing their views. The specific contents of the speeches are as follows: [slot\_2]. It is known that there are a total of [slot\_1] participants in this discussion. Question: Which among the [slot\_1] participants are telling the truth? Please list the names separated by commas, and in the order of their speeches, for example: Amelia, Bella, Taylor."
    \item "In a class-themed group discussion, [slot\_1] members expressed their opinions in sequence, with the specific content being: [slot\_2]. So who among these [slot\_1] members is telling the truth? Please separate the names with commas and list them in the order they spoke, for example: Charlie, Lily, Alexander."
    \item "During the weekend, members of the photography enthusiasts group gathered together to share their shooting experiences, and each person expressed their own insights on photography one by one. The specific content is as follows: [slot\_2]. It is known that a total of [slot\_1] people participated in this exchange. Question: Among these [slot\_1] people, who are the ones telling the truth? Please list the names separated by commas in the order they spoke, for example: Lee, Tom, Jerry"
    \item "Observe the chat records of a certain study group. There are [slot\_1] members who sent messages one after another, and the content is: [slot\_2]. Question: Among these [slot\_1] members who sent messages, who are the ones telling the truth? Please list the names separated by commas in the order of speaking, for example: William, James, John"
    \item "In the debate competition organized by the school, there are [slot\_1] debaters on each side who take turns presenting their viewpoints. The specific debate speeches are as follows: [slot\_2]. Question: Among these [slot\_1] debaters, who told the truth? Please list the names separated by a comma and in the order of speaking, for example: Michael, David, Liam."
    \item "At the family gathering, family members discussed weekend travel plans. Everyone took turns sharing their thoughts, specifically: [slot\_2]. It is known that there are [slot\_1] family members participating in the discussion. The question is: Who among these [slot\_1] people is telling the truth? Please list the names in the order they spoke, separated by commas, e.g., Noah, Oliver, Mary."
    \item "The community volunteer group held a meeting to plan next weekend's activity. The group members expressed their opinions on the activity plan one by one, and the specific speeches are as follows: [slot\_2]. It is known that there are [slot\_1] volunteers participating in this meeting. Question: Among these [slot\_1] people, who are telling the truth? Please list their names in the speaking order separated by commas, for example: John, Jennifer, Jessica"
    \item "Members of a certain book club gathered to discuss a popular novel. Each person shared their understanding of the novel's ending in turn, as follows: [slot\_2]. It is known that there are [slot\_1] participants in this book club. The question is: among these [slot\_1] members, who is telling the truth? Please list the names separated by commas and in the order they spoke, for example: Emma, Harry, Sophia"
    \item "At the company's quarterly goal kick-off meeting, team members successively presented their feasibility analyses for achieving the goals. The specific speeches were: [slot\_2]. It is known that there are a total of [slot\_1] project team members attending. Question: Among these [slot\_1] people, who are telling the truth? Please list the names separated by commas and in the order of speaking, for example: Aaron, Michael, Mary"
    \item "On a popular forum, there are [slot\_1] users discussing 'whether new energy vehicles should be promoted'. The original poster initiated the topic, and then other users responded one by one to express their views. The specific content is as follows: [slot\_2]. Question: Among the [slot\_1] users participating in the discussion, who is telling the truth? Please list the names separated by commas in the order of speaking, for example: David, Jennifer, Abel"
\end{enumerate} \\
\midrule
\rowcolor{gray!10} \textit{Input Code} & \fontsize{7pt}{8pt}\selectfont \texttt{def input(difficulty: int = 1, language: str = 'en'):} \\
\rowcolor{gray!10} & \fontsize{7pt}{8pt}\selectfont \texttt{~~import random, re; common\_surnames = []} \\
\rowcolor{gray!10} & \fontsize{7pt}{8pt}\selectfont \texttt{~~if language == 'en':} \\
\rowcolor{gray!10} & \fontsize{7pt}{8pt}\selectfont \texttt{~~~~common\_surnames = ['Smith', 'Johnson', 'Williams', 'Brown', 'Jones', 'Garcia', 'Miller', 'Davis', 'Rodriguez', 'Martinez', 'Hernandez', 'Lopez', 'Gonzalez', 'Wilson', 'Anderson', 'Thomas', 'Taylor', 'Moore', 'Jackson', 'Martin', 'Lee', 'Perez', 'Thompson', 'White', 'Harris', 'Sanchez', 'Clark', 'Ramirez', 'Lewis', 'Robinson', 'Walker', 'Young', 'Allen', 'King', 'Wright', 'Scott', 'Torres', 'Nguyen', 'Hill', 'Flores', 'Green', 'Adams', 'Nelson', 'Baker', 'Hall', 'Rivera', 'Campbell', 'Mitchell', 'Carter', 'Roberts', 'Gomez', 'Phillips', 'Evans', 'Turner', 'Diaz', 'Parker', 'Cruz', 'Edwards', 'Collins', 'Reyes', 'Stewart', 'Morris', 'Morales', 'Murphy', 'Cook', 'Rogers', 'Gutierrez', 'Ortiz', 'Morgan', 'Cooper', 'Peterson', 'Bailey', 'Reed', 'Kelly', 'Howard', 'Ramos', 'Kim', 'Cox', 'Ward', 'Richardson', 'Watson', 'Brooks', 'Chavez', 'Wood', 'James', 'Bennett', 'Gray', 'Mendoza', 'Ruiz', 'Hughes', 'Price', 'Alvarez', 'Castillo', 'Sanders', 'Patel', 'Myers', 'Long', 'Ross', 'Foster', 'Jimenez']} \\
\rowcolor{gray!10} & \fontsize{7pt}{8pt}\selectfont \texttt{~~def generate\_unique\_solution\_problem():} \\
\rowcolor{gray!10} & \fontsize{7pt}{8pt}\selectfont \texttt{~~~~while True:} \\
\rowcolor{gray!10} & \fontsize{7pt}{8pt}\selectfont \texttt{~~~~~~names, statements = generate\_random\_truth\_falsehood\_problem(difficulty)} \\
\rowcolor{gray!10} & \fontsize{7pt}{8pt}\selectfont \texttt{~~~~~~solutions = solve\_truth\_statements(names, statements)} \\
\rowcolor{gray!10} & \fontsize{7pt}{8pt}\selectfont \texttt{~~~~~~if len(solutions) == 1:} \\
\rowcolor{gray!10} & \fontsize{7pt}{8pt}\selectfont \texttt{~~~~~~~~true\_speakers = [name for name, truth in solutions[0].items() if truth == 'Truth']} \\
\rowcolor{gray!10} & \fontsize{7pt}{8pt}\selectfont \texttt{~~~~~~~~return names, statements, solutions[0], len(true\_speakers), true\_speakers} \\
\rowcolor{gray!10} & \fontsize{7pt}{8pt}\selectfont \texttt{~~def generate\_random\_truth\_falsehood\_problem(difficulty):} \\
\rowcolor{gray!10} & \fontsize{7pt}{8pt}\selectfont \texttt{~~~~difficulty\_params = \{1: \{"people\_count": 7\}, 2: \{"people\_count": 9\}, 3: \{"people\_count": 11\}, 4: \{"people\_count": 12\}, 5: \{"people\_count": 13\}, 6: \{"people\_count": 14\}, 7: \{"people\_count": 15\}, 8: \{"people\_count": 16\}, 9: \{"people\_count": 18\}, 10: \{"people\_count": 20\}\}} \\
\rowcolor{gray!10} & \fontsize{7pt}{8pt}\selectfont \texttt{~~~~count = difficulty\_params[difficulty]["people\_count"]} \\
\rowcolor{gray!10} & \fontsize{7pt}{8pt}\selectfont \texttt{~~~~surnames = random.sample(common\_surnames, count); names = [f"\{s\}" for s in surnames]} \\
\rowcolor{gray!10} & \fontsize{7pt}{8pt}\selectfont \texttt{~~~~statements = []; used = set()} \\
\rowcolor{gray!10} & \fontsize{7pt}{8pt}\selectfont \texttt{~~~~while len(statements) < count:} \\
\rowcolor{gray!10} & \fontsize{7pt}{8pt}\selectfont \texttt{~~~~~~mode = random.choices(["at least", "at most", "exactly"], [4, 1, 3])[0]} \\
\rowcolor{gray!10} & \fontsize{7pt}{8pt}\selectfont \texttt{~~~~~~target = random.choice(["truth", "lie"]); num = random.randint(1, count)} \\
\rowcolor{gray!10} & \fontsize{7pt}{8pt}\selectfont \texttt{~~~~~~stmt = f"There are \{mode\} \{num\} people telling the \{target\}."} \\
\rowcolor{gray!10} & \fontsize{7pt}{8pt}\selectfont \texttt{~~~~~~if stmt not in used: used.add(stmt); statements.append(stmt)} \\
\rowcolor{gray!10} & \fontsize{7pt}{8pt}\selectfont \texttt{~~~~return names, statements} \\
\rowcolor{gray!10} & \fontsize{7pt}{8pt}\selectfont \texttt{~~def solve\_truth\_statements(names, stmts):} \\
\rowcolor{gray!10} & \fontsize{7pt}{8pt}\selectfont \texttt{~~~~n = len(names); result = []; pattern = r"There are (at least|at most|exactly) (\textbackslash\textbackslash{}d+) people telling the (truth|lie)"} \\
\rowcolor{gray!10} & \fontsize{7pt}{8pt}\selectfont \texttt{~~~~from itertools import product} \\
\rowcolor{gray!10} & \fontsize{7pt}{8pt}\selectfont \texttt{~~~~for possibility in product([True, False], repeat=n):} \\
\rowcolor{gray!10} & \fontsize{7pt}{8pt}\selectfont \texttt{~~~~~~true\_count = sum(possibility); false\_count = n - true\_count; checks = []} \\
\rowcolor{gray!10} & \fontsize{7pt}{8pt}\selectfont \texttt{~~~~~~for stmt in stmts:} \\
\rowcolor{gray!10} & \fontsize{7pt}{8pt}\selectfont \texttt{~~~~~~~~match = re.match(pattern, stmt); m\_str, num, k\_str = match.groups()} \\
\rowcolor{gray!10} & \fontsize{7pt}{8pt}\selectfont \texttt{~~~~~~~~curr = true\_count if k\_str == 'truth' else false\_count; num = int(num)} \\
\rowcolor{gray!10} & \fontsize{7pt}{8pt}\selectfont \texttt{~~~~~~~~if m\_str == 'at least': checks.append(curr >= num)} \\
\rowcolor{gray!10} & \fontsize{7pt}{8pt}\selectfont \texttt{~~~~~~~~elif m\_str == 'at most': checks.append(curr <= num)} \\
\rowcolor{gray!10} & \fontsize{7pt}{8pt}\selectfont \texttt{~~~~~~~~else: checks.append(curr == num)} \\
\rowcolor{gray!10} & \fontsize{7pt}{8pt}\selectfont \texttt{~~~~~~valid = True} \\
\rowcolor{gray!10} & \fontsize{7pt}{8pt}\selectfont \texttt{~~~~~~for i in range(n):} \\
\rowcolor{gray!10} & \fontsize{7pt}{8pt}\selectfont \texttt{~~~~~~~~if possibility[i] != checks[i]: valid = False; break} \\
\rowcolor{gray!10} & \fontsize{7pt}{8pt}\selectfont \texttt{~~~~~~if valid: result.append(\{names[j]: 'Truth' if possibility[j] else 'Lie' for j in range(n)\})} \\
\rowcolor{gray!10} & \fontsize{7pt}{8pt}\selectfont \texttt{~~~~return result} \\
\rowcolor{gray!10} & \fontsize{7pt}{8pt}\selectfont \texttt{~~names, stmts, assign, t\_cnt, t\_spks = generate\_unique\_solution\_problem()} \\
\rowcolor{gray!10} & \fontsize{7pt}{8pt}\selectfont \texttt{~~process = " ".join([f"\{n\}: \{s\}" for n, s in zip(names, stmts)])} \\
\rowcolor{gray!10} & \fontsize{7pt}{8pt}\selectfont \texttt{~~return ', '.join(t\_spks), [len(names), process]} \\
\midrule
\rowcolor{white} \textit{Solution Code} & \fontsize{7pt}{8pt}\selectfont \texttt{def solution(params, language: str = 'en'): return params} \\
\midrule
\rowcolor{gray!10} \textit{BFR Type} & F1-score \\
\midrule
\rowcolor{white} \textit{BFR Code} & \fontsize{7pt}{8pt}\selectfont \texttt{def f1\_score\_reward(model\_answer, ground\_truth):} \\
\rowcolor{white} & \fontsize{7pt}{8pt}\selectfont \texttt{~~gt\_set = set(ground\_truth.split(', '))} \\
\rowcolor{white} & \fontsize{7pt}{8pt}\selectfont \texttt{~~model\_set = set(model\_answer.split(', '))} \\
\rowcolor{white} & \fontsize{7pt}{8pt}\selectfont \texttt{~~if not gt\_set or not model\_set: return -1.0} \\
\rowcolor{white} & \fontsize{7pt}{8pt}\selectfont \texttt{~~inter = len(gt\_set.intersection(model\_set))} \\
\rowcolor{white} & \fontsize{7pt}{8pt}\selectfont \texttt{~~p = inter / len(model\_set); r = inter / len(gt\_set)} \\
\rowcolor{white} & \fontsize{7pt}{8pt}\selectfont \texttt{~~f1 = 2 * (p * r) / (p + r) if (p + r) > 0 else 0} \\
\rowcolor{white} & \fontsize{7pt}{8pt}\selectfont \texttt{~~return 1.0 if f1 == 1.0 else f1 - 1.0} \\
\end{longtable}
\endgroup

\twocolumn
\section{Detailed Experiment Results}
\label{sec:appendix_exp_details}

This appendix details the experiment results. Based on the verl\footnote{\url{https://github.com/volcengine/verl}} framework, 64 GPUs were utilized for GRPO training. The training of Qwen3-8B took approximately 120 hours for 2 epochs, while the training of Qwen3-14B consumed around 300 hours for the same number of epochs\footnote{\url{https://huggingface.co/collections/Qwen/qwen3}}.
\subsection{Difficulty Matching Ablation Study} 
\label{Difficulty Matching Ablation Study}

\begin{figure*}[t]
    \centering
    \includegraphics[width=\textwidth]{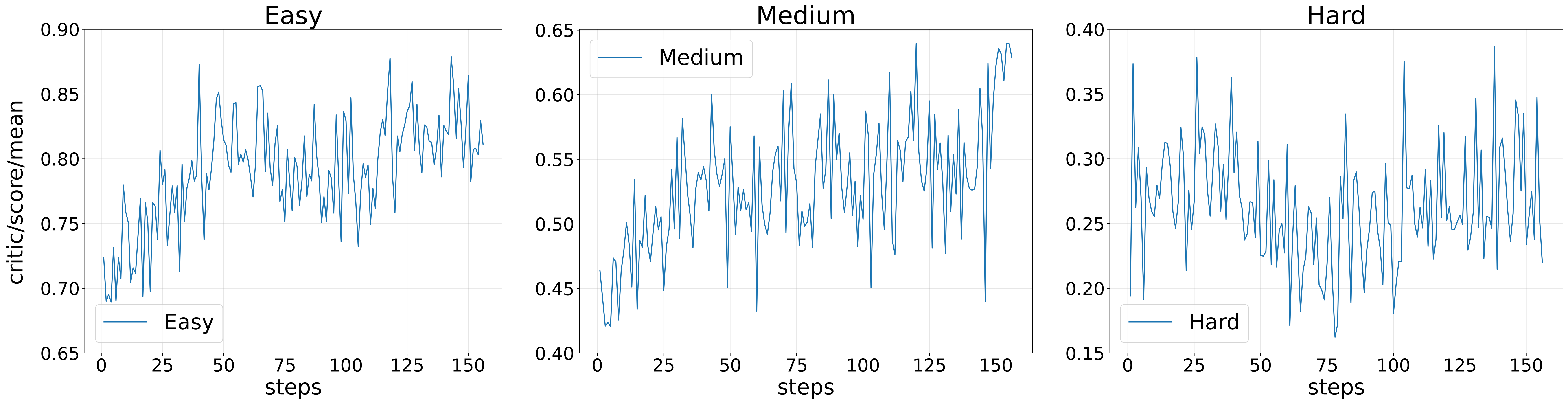}
    \caption{The critic/score/mean metrics of Qwen3-8B during the GRPO process on the different training sets of varying difficulty levels, including Easy, Medium, and Hard.}
    \label{fig:qwen3_8B_easy_medium_hard_train_score}
\end{figure*}

\begin{figure*}[t]
    \centering
    \includegraphics[width=\textwidth]{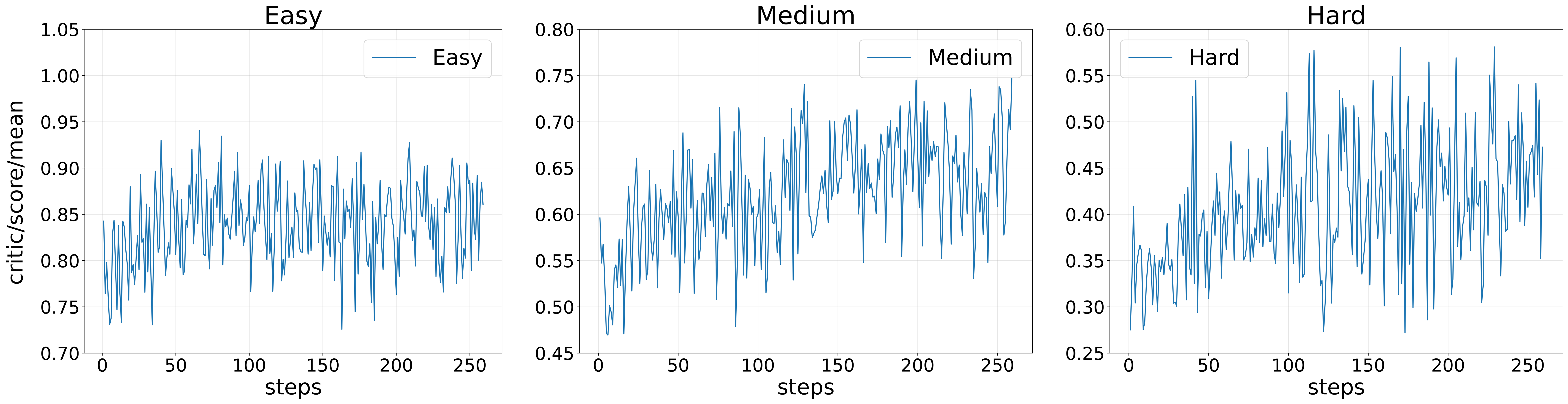}
    \caption{The critic/score/mean metrics of Qwen3-14B during the GRPO process on the different training sets of varying difficulty levels, including Easy, Medium, and Hard.}
    \label{fig:qwen3_14B_easy_medium_hard_train_score}
\end{figure*}

As shown in the~\Cref{fig:qwen3_8B_easy_medium_hard_train_score} and~\Cref{fig:qwen3_14B_easy_medium_hard_train_score}, they illustrate the critic/score/mean metrics during the GRPO process over 2 training epochs for Qwen3-8B and Qwen3-14B on the different training datasets with varying difficulty levels, including Easy, Medium, and Hard.

As shown in \Cref{fig:qwen3_8B_easy_medium_hard_train_score}, the increase in critic/score/mean is particularly pronounced for Qwen3-8B when training on Easy difficulty data, followed by Medium, and finally Hard. Although the curve for Easy shows some fluctuations, it maintains an upward trend throughout the later stages of training; whereas the curve for Medium rises significantly during the early stages of training but experiences greater oscillations in the later stages; the curve for Hard is characterized by substantial oscillations overall, with no significant increase at the end of the entire training period compared to the beginning.

As shown in \Cref{fig:qwen3_14B_easy_medium_hard_train_score}, when training with Medium difficulty data, the critic/score/mean of Qwen3-14B increased particularly significantly, followed by Hard, and finally Easy. Although the curve for Easy showed an upward trend in the early stages, it quickly approached convergence; whereas the curve for Medium consistently exhibited a clear upward trend with smaller oscillations; the curve for Hard showed an overall slight upward trend, but with more pronounced oscillations in the second epoch.

Based on the experimental phenomena observed above, it can be generally concluded that when the model is trained on relatively simple data, its performance can significantly improve. However, overly simple data can lead to rapid convergence, resulting in insignificant performance enhancement. Conversely, when the model is trained on more difficult data, its performance exhibits significant fluctuations, which may even lead to training collapse. Therefore, in GRPO training, selecting the most suitable data for the model's adaptability to difficulty is particularly crucial, and this point is highly consistent with the performance on the evaluation set shown in \Cref{tab:difficulty_ablation_appendix}.

\begin{table}[!ht]
\centering
\normalsize
\renewcommand{\arraystretch}{1.3}
\setlength{\tabcolsep}{3pt}
\resizebox{\linewidth}{!}{
\begin{tabular}{lcccccc}
\toprule
\textbf{Model} & \textbf{AIME24} & \textbf{AIME25} & \textbf{HMMT25} & \textbf{BBH} & \textbf{BBEH} & \textbf{ARC-AGI} \\
\midrule
Qwen3-8B  & 75.2 & 66.1 & 47.1 & 88.2 & 29.2 & 4.0 \\
Easy-1epoch & \underline{78.3} & 67.5 &	50.3 &	\underline{89.3} &	\underline{30.4} &	\underline{4.3} \\
\textbf{Easy-2epoch}      & \textbf{81.7} & \textbf{69.1} & \textbf{52.3} & \textbf{90.2} & \textbf{31.1} & \textbf{4.6} \\
Medium-1epoch & 77.5 &	66.7 &	50.1 &	88.5 &	30.2 &	4.1 \\
Medium-2epoch    & 77.6 & \underline{67.6} & \underline{50.9} & 88.7 & 29.6 & 4.2 \\
Hard-1epoch & 75.8 &	64.5 &	48.4 &	88.0 &	28.8 &	3.9 \\
Hard-2epoch      & 76.5 & 65.1 & 50.0 & 88.3 & 29.0 & 4.0 \\
\midrule
Qwen3-14B & 76.7 & 70.3 & 54.3 & 89.6 & 32.2 & 6.3 \\
Easy-1epoch & 78.3 &	71.6 &	56.9 &	90.9 &	33.6 &	6.3 \\
Easy-2epoch      & 78.8 & 72.5 & 58.6 & 90.9 & 33.7 & 6.1 \\
Medium-1epoch & \underline{81.6} &	\underline{74.2} &	\underline{61.5} &	\underline{91.4} &	\underline{35.6} &	\underline{7.6} \\
\textbf{Medium-2epoch}    & \textbf{82.4} & \textbf{75.8} & \textbf{63.1} & \textbf{92.1} & \textbf{36.7} & \textbf{7.9} \\
Hard-1epoch & 80.0 &	72.9 &	59.5 &	90.8 &	34.1 &	6.9 \\
Hard-2epoch      & 80.8 &	73.8 &	59.9 &	90.9 &	34.7   &	6.8  \\
\bottomrule
\end{tabular}
}
\caption{Ablation study on model scale and task difficulty.}
\label{tab:difficulty_ablation_appendix}
\end{table}

\subsection{Bipolar Float Reward Ablation Study}
\label{Bipolar Float Reward Ablation Study}

\begin{figure*}[t]
    \centering
    \includegraphics[width=\textwidth]{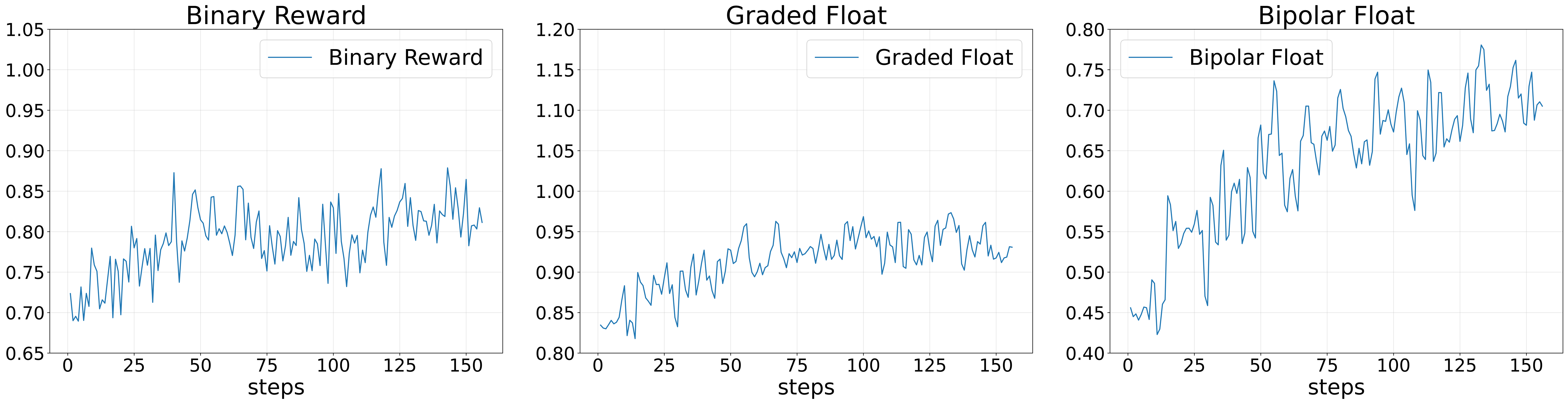}
    \caption{The critic/score/mean metrics of Qwen3-8B during the GRPO process by using different reward mechanisms, including Binary Reward, Graded Float Reward, and Bipolar Float Reward.}
    \label{fig:qwen3_8B_float_reward_train_score}
\end{figure*}

\Cref{fig:qwen3_8B_float_reward_train_score} illustrates the critic/score/mean metrics during the GRPO process over 2 training epochs for Qwen3-8B by using different reward mechanisms, including Binary Reward, Graded Float Reward, and Bipolar Float Reward. From the curves in the figure, although the Bipolar Float Reward shows somewhat lower overall values due to the negative score component compared to Binary Reward, its curve demonstrates a more smoothly and steadily increasing trajectory. In contrast, the Graded Float Reward, which consists entirely of positive scores, achieves the highest curve values, but demonstrates only minimal overall growth, showing minimal distinction from Binary Reward. This phenomenon is highly consistent with the evaluation results presented in \Cref{tab:reward_ablation_appendix}.

\begin{table}[h]
\centering
\normalsize
\renewcommand{\arraystretch}{1.3}
\setlength{\tabcolsep}{3pt}
\resizebox{\linewidth}{!}{
\begin{tabular}{lcccccc}
\toprule
\textbf{Reward} & \textbf{AIME24} & \textbf{AIME25} & \textbf{HMMT25} & \textbf{BBH} & \textbf{BBEH} & \textbf{ARC-AGI} \\
\midrule
Qwen3-8B & 75.2 & 66.1 & 47.1 & 88.2 & 29.2 & 4.0 \\
Binary Reward-1epoch & 78.3 &	67.5 &	50.3 &	89.3 &	30.4 &	4.3 \\
Binary Reward-2epoch & 81.7 & 69.1 & 52.3 & 90.2 & \underline{31.1} & \underline{4.6} \\
Graded Float-1epoch & 76.4 &	66.2 &	53.6 &	88.9 &	30.2 &	4.2 \\
Graded Float-2epoch & 76.9 &	66.3 &	53.0 &	\underline{90.4} &	31.0 &	4.3 \\
Bipolar Float-1epoch & \underline{81.8} &	\underline{69.2} &	\underline{54.9} &	89.7 &	\underline{31.1} &	\underline{4.6} \\
\textbf{Bipolar Float-2epoch} & \textbf{82.6} &	\textbf{71.3} &	\textbf{56.6} &	\textbf{91.1} &	\textbf{32.5} &	\textbf{4.7} \\

\bottomrule
\end{tabular}
}
\caption{Ablation study on reward mechanisms. Comparison of Binary, Graded Float, and Bipolar Float rewards.}
\label{tab:reward_ablation_appendix}
\end{table}

\clearpage
\twocolumn

\raggedbottom
\section{Prompts}
\label{sec:appendix_prompts}
This appendix presents the specific prompt templates used in the \textsc{UltraLogic} framework.

\subsection{The Prompt of Task Mining and Template Creation}
\label{The Prompt of Task Mining and Template Creation}
This prompt acts as an initial gatekeeper and architect, determining whether a reasoning problem can be programmatically generated and, if so, transforming it into a natural language logical reasoning template with placeholders (slots).

\begin{tcolorbox}[breakable, after skip=3em, colback=mintgreen, width=\columnwidth, colframe=black!50, title=Task Mining and Template Creation Prompt]
\small
You are an intelligent assistant for task mining and question template generation. Your task is to judge whether a given reasoning problem can be solved by code and whether its input can be programmatically generated. If feasible, rewrite it into a daily life logical reasoning template.

\textbf{Requirements:}
\begin{enumerate}
    \item \textbf{Feasibility Check:} If the problem cannot be solved by code or the inputs cannot be generated, reply "Unable to generate a code-solvable template."
    \item \textbf{Scenario Rewriting:} Convert the problem into a natural daily life reasoning question by introducing backgrounds, names, and locations.
    \item \textbf{Slot Replacement:} Replace specific input parameters with [Slot 1], [Slot 2], etc. Ensure the required answer logic remains identical to the original problem.
    \item \textbf{Technical Constraint:} Do not include any code snippets or technical terms (e.g., arrays, hash maps, binary trees).
    \item \textbf{Output Format:} If feasible, output the result in JSON format with three keys: \texttt{rewriting\_plan}, \texttt{new\_question\_template}, and \texttt{slot\_description}.
\end{enumerate}
\textbf{Example:} \\
\texttt{[Reasoning Problem]} Walking in the park, if you
follow these routes.. \\
\texttt{[Output]} Unable to generate a code-solvable template. \\
or \\
\begin{verbatim}
{
    "rewriting_plan": "xxx",
    "new_question_template": "xxx",
    "slot_description": "xxx"
}
\end{verbatim}

\textbf{Input:} \\
\texttt{[Reasoning Problem]} \{reasoning\_problem\} \\
\texttt{[Output]} 
\end{tcolorbox}

\subsection{The Prompt of Question Template Generalization}
\label{The Prompt of Question Template Generalization}
This prompt is used to expand the diversity of the dataset by rewriting a single logical template into ten different scenarios while maintaining the underlying logic and slot structure.

\begin{tcolorbox}[breakable, after skip=3em, colback=mintgreen, width=\columnwidth, colframe=black!50, title=Question Template Generalization Prompt]
\small
You are a rewriting generalizer. Your task is to rewrite a given question template into ten different versions. 
\begin{enumerate}
    \item Randomly introduce backgrounds, change scenarios, names, and word orders to ensure diversity.
    \item Maintain the same question type and core logic; do not change the intent of the test.
    \item Ensure that the slots (e.g., [Slot 1], [Slot 2]) remain unchanged in the new templates.
    \item The similarity between the ten generated templates must be low; do not simply swap names.
    \item Output in JSONArray format with two keys: \texttt{rewriting\_plan}, \texttt{new\_question\_template}.
    \item Maintain the original language of the template.
\end{enumerate}

\textbf{Example:} \\
\texttt{[Original Template]} Walking in the park, if you follow these routes... [Slot 1] \\
\texttt{[Output JSONArray]} 
(Ten variations including maze exploration, robot navigation, etc.)
\begin{verbatim}
[
    {
        "rewriting_plan": "xxx",
         "new_question_template": "xxx"
    },
    ...
]
\end{verbatim}

\textbf{Task:} \\
Rewrite the following template into ten variations: \\
\texttt{[Original Template]} \{question\_template\} \\
\texttt{[Output JSONArray]} 
\end{tcolorbox}

\subsection{The Prompt of Input Code Generation}
\label{The Prompt of Input Code Generation}
The following prompt is used to guide the model to generate Python scripts. these scripts are responsible for producing randomized, difficulty-controlled input text that fits into the predefined question templates.

\begin{tcolorbox}[breakable, after skip=3em, colback=mintgreen, width=\columnwidth, colframe=black!50, title=Input Code Generation Prompt]
\small
You are a code generator. Please generate input generation code based on a given original programming problem, logic reasoning template, slot description, and input examples. Requirements are as follows:
\begin{enumerate}
    \item The function of the code: According to the parameters in the original problem, generate natural language text for various slots based on the examples. This ensures the template forms a complete, logical, and easy-to-understand reasoning problem.
    \item Input parameter: \texttt{difficulty} (1--10). It controls the complexity so that different levels of difficulty are covered and the span between levels is significant. In the input code, there will be a mapping of difficulty mapping parameters, for example:
\begin{verbatim}
difficulty_params = {
    1: {"num": 3}, 2: {"num": 5},
    3: {"num": 8}, 4: {"num": 10},
    5: {"num": 12}, 6: {"num": 15},
    7: {"num": 18}, 8: {"num": 22},
    9: {"num": 26}, 10: {"num": 30}
}
\end{verbatim}
    \item Input parameter: \texttt{language} (en, zh). It controls the language of the slot-filling text, whether it is in Chinese or English.
    \item The code should be concise, clear, and handle edge cases.
    \item The generated text must be randomized. Do not output raw data formats like arrays, lists, or JSON; convert them into natural language descriptions.
    \item The return value should contain two elements: the raw parameter values (as in the example) and a list of strings for the slots.
    \item Ensure generated values are logically consistent (e.g., no "893:00" in time contexts) and the distribution of answers is as uniform as possible.
\end{enumerate}

\textbf{Example:} \\
\texttt{[Question Template]} There are [Slot 1] members who sent messages one after another... \\
\texttt{[Slot Desc]} [Slot 1] is number of people... \\
\texttt{[Input Code]} (Generated Python function \texttt{input(difficulty, language='en')})

\textbf{Task:} \\
Generate the code for the following: \\
\texttt{[Template]} \{question\_template\} \\
\texttt{[Slot Desc]} \{slot\_description\} \\
\texttt{[Input Code]}
\end{tcolorbox}

\subsection{The Prompt of Solution code Generation}
\label{The Prompt of Solution code Generation}
This prompt guides the model to generate the core solver script. This Python code takes the randomized slot values as input and computes the definitive ground-truth answer according to the problem's logic.

\begin{tcolorbox}[breakable, after skip=3em, colback=mintgreen, width=\columnwidth, colframe=black!50, title=Solution Code Generation Prompt]
\small
You are a Python expert. Your task is to write a solver function for a logical reasoning problem based on its original programming logic and template.

\textbf{Requirements:} 
\begin{enumerate}[leftmargin=*, nosep] 
    \item \textbf{Core Function}: \texttt{def solution(input\_params, language)} that takes raw parameter values returned by input code as input\_parameters and returns the correct answer. 
    \item \textbf{Logic Accuracy}: The code must strictly implement the reasoning logic described in the \texttt{[Programming Problem]}. 
    \item \textbf{Code Style}: Ensure the code is concise, efficient, and includes necessary comments for complex steps. 
    \item \textbf{Output Format}: The function should return a single value (string, int, or float) that can be directly compared with model outputs. 
\end{enumerate}

\textbf{Example:} \\
\texttt{[Question Template]} There are [Slot 1] members who sent messages one after another... \\
\texttt{[Slot Desc]} [Slot 1] is number of people... \\
\texttt{[Complete Question]} There are 5 members who sent messages one after another... \\
\texttt{[Input Parameters]} [5] \\
\texttt{[Solution Code]} (Generated Python function \texttt{solution(input\_params, language='en')})

\textbf{Task:} \\
Generate the Python \texttt{solution} function for this problem. \\
\texttt{[Question Template]} \{question\_template\} \\
\texttt{[Slot Desc]} \{slot\_desc\} \\
\texttt{[Complete Question]} \{complete\_question\} \\
\texttt{[Input Parameters]} \{input\_parameters\} \\
\texttt{[Solution Code]}
\end{tcolorbox}

\subsection{The Prompt of Dynamic Parameter Adjustment}
\label{The Prompt of Dynamic Parameter Adjustment}
This prompt is used to determine and refine the mapping between difficulty levels ($1 \sim 10$) and internal complexity parameters. It ensures that structural complexity scales appropriately and can be adjusted based on the actual success rate of models.

\begin{tcolorbox}[breakable, after skip=3em, colback=mintgreen, width=\columnwidth, colframe=black!50, title=Dynamic Parameter Adjustment Prompt]
\small
You are a mathematical modeling assistant for logic puzzles. Your task is to define or adjust a mapping between a difficulty scale (1 to 10) and the structural parameters of a reasoning problem.

\textbf{Instructions:}
\begin{enumerate}[leftmargin=*, nosep]
    \item Identify core variables (e.g., $N$ people, $M$ steps).
    \item Create or update the \texttt{difficulty\_params} dictionary for \textbf{all levels from 1 to 10}.
    \item \textbf{Global Calibration}: If measured success rates for anchor levels (e.g., 1, 5, and 10) are provided, evaluate the overall difficulty trend.
    \item \textbf{Monotonic Consistency}: Ensure that parameters increase \textbf{monotonically and linearly} across the 1--10 range. If a middle level (e.g., Level 5) is adjusted, intermediate levels must be shifted accordingly to maintain a smooth gradient and avoid "difficulty inversion".
\end{enumerate}

\textbf{Example (Global Adjustment):} \\
\texttt{[Question Template]} There are [Slot 1] members who sent messages one after another... \\
\texttt{[Slot Desc]} [Slot 1] is number of people... \\
\texttt{[Current Mapping]}
\begin{verbatim}
difficulty_params = {
    1: {"num": 3}, 2: {"num": 6},
    3: {"num": 9}, 4: {"num": 12},
    5: {"num": 15}, 6: {"num": 18},
    7: {"num": 21}, 8: {"num": 24},
    9: {"num": 27}, 10: {"num": 30}
}
\end{verbatim}
\texttt{[Feedback]} \texttt{L1:100\%, L5:30\% (Target 50\%), L10:0\%}. \\
\texttt{[Adjustment Logic]} The model struggles as \texttt{num} exceeds 12. To maintain linear growth, we compress the parameters for levels 2--9. \\
\texttt{[Updated Mapping]} 
\begin{verbatim}
difficulty_params = {
    1: {"num": 3}, 2: {"num": 5},
    3: {"num": 8}, 4: {"num": 10},
    5: {"num": 12}, 6: {"num": 15},
    7: {"num": 18}, 8: {"num": 22},
    9: {"num": 26}, 10: {"num": 30}
}
\end{verbatim}

\textbf{Task:} \\
Suggest/Refine the complete 1--10 parameter mapping based on the following feedback: \\
\texttt{[Question Template]} \{question\_template\} \\
\texttt{[Slot Desc]} \{slot\_desc\} \\
\texttt{[Current Mapping]} \{currect\_mapping\} \\
\texttt{[Feedback]} \{feedback\} \\
\texttt{[Adjustment Logic]}\\
\texttt{[Updated Mapping]}
\end{tcolorbox}

\section{Details of Human Annotation}
\label{Details of Human Annotation}

In our study, a total of 12 annotators(bachelor's degree or above in computer-related majors) participated in task logic verification, template refinement, and reward mechanism configuration; all of them were internal members of our research team.
Here presents the comprehensive guidelines provided to human annotators.

\begin{tcolorbox}[breakable, colback=blue!5, colframe=black!50, title=Guidelines for Human Annotators]

\small
\textbf{(1) Question Quality Check}
\begin{itemize}[leftmargin=*, nosep]
    \item \textbf{Ambiguity Analysis}: Check for logic loopholes, multiple possible interpretations, missing/redundant information, unstated implicit assumptions, misleading statements, or vague phrasing (e.g., unclear temporal/spatial relationships).
    \item \textbf{Judgment Criteria}: A task is valid only if it has a unique reasonable answer, presents all necessary information for solving, and contains no presets that violate common sense.
\end{itemize}

\textbf{(2) Category Labeling}: Annotate each task according to the descriptions of the \textit{Three-dimensional Orthogonal Classification System} (Task Domain, Core Ability, and Difficulty Source).

\textbf{(3) Template Verification}
\begin{itemize}[leftmargin=*, nosep]
    \item Ensure slots are compatible with the new templates and no critical information is lost.
    \item Verify that the 10 generated templates are visually distinct and that the core reasoning focal point has not been altered by rephrasing or scenario changes.
\end{itemize}

\textbf{(4) Difficulty and Correctness Verification}: Validate the task through multiple sampling of LLMs. Ensure that responses are complete and scoring is accurate, ruling out "noise" such as low success rates caused by API call failures.

\textbf{(5) Input and Solution Code Verification}
\begin{itemize}[leftmargin=*, nosep]
    \item \textbf{Manual Correction}: If full automation fails, annotators must manually intervene to ensure code correctness based on the principle of accurate parameter generation and solution logic.
    \item \textbf{Difficulty Adjustment Principles}:
    \begin{enumerate}[label=\roman*., leftmargin=*, nosep]
        \item \textit{Complexity Scaling}: Increase distractors (recommended ratio of 3:7 for core vs. distractor info), design multi-step chains (3--5 logical jumps), or hide key clues in secondary descriptions.
        \item \textit{Advanced Techniques}: Utilize nested structures (e.g., contradictory testimonies), psychological misdirection (exploiting cognitive biases), or domain-specific knowledge (e.g., cryptography).
        \item \textit{Optimization}: Maintain strict logical rigor and ensure all clues are presented fairly and are traceable.
    \end{enumerate}
\end{itemize}

\textbf{(6) Bipolar Float Reward (BFR) Annotation}
\begin{itemize}[leftmargin=*, nosep]
    \item \textbf{Response Restructuring}: Convert single answers into structured combinations that reflect the reasoning process.
    \item \textbf{Scoring Configuration}: Select from standard types: Accuracy, F1-Score, Similarity, or Absolute Difference.
    \item \textbf{Robustness}: The scoring code must handle minor output instabilities, such as redundant spaces or varied punctuation (bilingual compatibility).
    \item \textbf{Determinism}: Functions must yield identical results for the same input with no randomness. Returns must be a float in $[-1, 0) \cup \{1\}$.
    \item \textbf{Difficulty Preservation}: Modifications should focus on output format (e.g., "list all IDs" instead of "count the IDs") without changing the problem's underlying constraints.
\end{itemize}

\end{tcolorbox}

\clearpage
\twocolumn

\end{document}